\documentclass{iopjournal}
\usepackage{cite}
\usepackage{amsmath,amssymb,amsfonts}
\usepackage{algorithmic}
\usepackage{graphicx}
\usepackage{textcomp}
\usepackage{wrapfig}
\usepackage{xcolor}
\usepackage{subcaption} % ?????
\usepackage{multirow}
\usepackage{float}%??float????
\usepackage{booktabs}%????\toprule?\midrule?\bottomrule
\usepackage{stfloats} % ??????????
\usepackage{comment}
\usepackage{makecell} % 允许表头换行
\usepackage{tabularx} 
\usepackage{cleveref}
\crefname{figure}{Fig.}{Figs.}
\Crefname{figure}{Fig.}{Figs.}
\begin{document}

\articletype{Article type} %	 e.g. Paper, Letter, Topical Review...

\title{PUL-SLAM: Path-Uncertainty Co-Optimization with Lightweight Stagnation Detection for Efficient Robotic Exploration}

\author{Yizhen Yin$^1$, Dapeng Feng$^1$,Hongbo Chen$^{1,*}$ and Yuhua Qi$^1$}

\affil{$^1$School of Systems Science and Engineering, Sun Yat-sen University, Guangzhou510220, People’s Republic of China}

\affil{$^*$Author to whom any correspondence should be addressed.}

\email{chenhongbo@mail.sysu.edu.cn}


\begin{abstract}
Existing Active SLAM methodologies face issues such as slow exploration speed and suboptimal paths. To address these limitations, we propose a hybrid framework combining a Path-Uncertainty Co-Optimization Deep Reinforcement Learning framework and a Lightweight Stagnation Detection mechanism. The Path-Uncertainty Co-Optimization framework jointly optimizes path length and pose uncertainty through a dual-objective reward function, balancing exploration and exploitation. The Lightweight Stagnation Detection reduces redundant exploration through Lidar Static Anomaly Detection and Map Update Stagnation Detection, terminating episodes on low expansion rates. Experimental results show that compared with the frontier-based method and RRT method, our approach shortens exploration time by up to 65$\%$ and reduces path distance by up to 42$\%$, significantly improving exploration efficiency in complex environments while maintaining reliable map completeness. Ablation studies confirm that the collaborative mechanism accelerates training convergence. Empirical validation on a physical robotic platform demonstrates the algorithm’s practical applicability and its successful transferability from simulation to real-world environments.
\end{abstract}

\section{Introduction}
\label{sec:introduction}
Active Simultaneous Localization and Mapping (Active SLAM) requires robots to simultaneously perform three critical tasks in unknown environments: environmental mapping, self-localization, and exploration path planning, thereby enabling efficient environmental exploration\cite{placed2023survey}. This technology has proven indispensable in critical application scenarios such as disaster rescue\cite{niroui2019deep}, planetary exploration\cite{nahavandi2025comprehensive}, underground mine exploration\cite{wang2025multi,ackerman2022robots}, and infrastructure inspection\cite{zhou2023racer}, particularly in environments that are inaccessible or hazardous to humans, where fully autonomous robotic exploration systems can substantially improve task execution efficiency while minimizing personnel risks\cite{azpurua2023survey}. The fundamental challenge of Active SLAM lies in achieving a dynamic equilibrium between exploration (discovering new areas) and exploitation (revisiting known regions to reduce localization and mapping uncertainty): robots must rapidly discover new areas while simultaneously conducting sufficient exploration of known regions to minimize uncertainties in positioning and mapping. Unlike traditional SLAM approaches that focus exclusively on mapping and localization accuracy, Active SLAM introduces significant additional complexity through path planning and exploration strategy decision-making, thereby transforming the problem into a highly complex multi-objective optimization challenge.

Active SLAM methodologies exhibit significant diversity in exploration strategies. Traditional frontier-based methods \cite{yamauchi1997frontier,keidar2014efficient,wang2024exploration,saravanan2024fit}, random sampling techniques like RRT \cite{umari2017autonomous,faust2018prm,wu2019autonomous}, and information-theoretic approaches \cite{whaite1997autonomous,asgharivaskasi2023semantic,stachniss2005information,carrillo2012comparison} have long dominated exploration strategies in unknown environments. These methods feature intuitive implementation architectures and are relatively easy to deploy. However, they employ static decision-making strategies that cannot adapt to dynamic environmental factors such as environmental complexity and exploration progress, resulting in unstable performance in varied environments. Furthermore, they typically optimize only a single objective (frontier-based methods focus on coverage ratio, RRT emphasizes path feasibility, and information-theoretic approaches prioritize uncertainty reduction), while neglecting the multi-objective trade-offs involving exploration efficiency, energy consumption, and other factors. This limitation, combined with their lack of global perspective, often leads to redundant backtracking paths in cluttered spaces. Recent deep reinforcement learning (DRL) methods \cite{zhao2024autonomous,zhou2024indoor,zhao2024exploration,cao2023ariadne,chen2024lidar,alcalde2022slam,botteghi2021curiosity,placed2020deep,zhu2018deep,chaplot2020learning,zhao2025multirobot} have demonstrated considerable potential in autonomous exploration domains. These approaches enable agents to interact with the environment and continuously refine their exploration strategies based on reward feedback, thereby optimizing the decision-making process. However, the performance of these methods is highly dependent on the design of the reward function. A well-designed reward function can steer the agent toward an optimal policy. Conversely, an improper design may result in the agent learning suboptimal strategies that significantly diverge from efficient exploration paths.

Despite these diverse paradigms, current systems still struggle to achieve an optimal balance between exploration efficiency and mapping accuracy, often suffering from slow exploration speeds and suboptimal trajectories. Additionally, existing systems lack mechanisms to dynamically detect and correct inefficient exploration behaviors, causing robots to become trapped in local oscillations or redundant exploration patterns when encountering complex obstacles. To address these limitations, we propose an innovative dual-layer collaborative optimization framework that jointly models path optimization and uncertainty reduction, while introducing a lightweight stagnation detection mechanism to enhance the system's adaptability in complex environments. The main contributions of this study include:
\begin{itemize}
\item \textbf{Path-Uncertainty Co-Optimization DRL Framework}: We propose a novel deep reinforcement learning framework that jointly optimizes travel distance and pose uncertainty through a dual-objective reward function, balancing exploration and exploitation. 
\item \textbf{Lightweight Stagnation Detection}: A Lightweight Stagnation Detection module (LSD) mitigates redundant exploration via real-time LiDAR analysis. Simultaneously, map-update detection terminates episodes on low expansion rates. This dual strategy reduces inefficiencies and suppresses learning-hindering behaviors.
\item \textbf{Extensive simulation and real-world experiments}: Experimental results show that compared with the frontier-based method and RRT method, the time is shortened by up to 65\%, and the path is shortened by up to 42\%, which significantly improves the exploration efficiency.
\end{itemize}

\section{RELATED WORK}
\subsection{Traditional Active SLAM}
Traditional Active SLAM methodologies can be broadly categorized into frontier-based exploration, random sampling techniques, and information-theoretic approaches, each with distinct characteristics and limitations that inform our current research direction.

Frontier-based exploration, pioneered by Yamauchi \cite{yamauchi1997frontier}, represents one of the most influential paradigms in robotic exploration. This approach identifies boundary regions between known and unknown areas (frontiers) and directs robots toward these locations to maximize information gain. Enhanced variants like Wavefront Frontier Detection (WFD) \cite{keidar2014efficient} improved computational efficiency but still suffer from significant drawbacks in complex environments: the algorithm tends to generate suboptimal paths with excessive backtracking, struggles with large-scale environments due to growing computational overhead, and lacks mechanisms for global trajectory optimization. Despite these limitations, frontier-based methods remain widely adopted in recent works \cite{wang2024exploration,saravanan2024fit} due to their intuitive implementation and reliable coverage performance.

Random sampling-based methods, particularly Rapidly-exploring Random Trees (RRT) \cite{umari2017autonomous}, offer an alternative exploration strategy by constructing search trees through random sampling of the configuration space. While RRT variants excel at finding feasible paths in high-dimensional spaces and complex obstacle arrangements, they exhibit critical shortcomings for exploration tasks: the resulting trajectories are often tortuous and energy-inefficient, coverage completeness is compromised due to undersampling in narrow passages, and the stochastic nature of sampling leads to inconsistent exploration patterns. These limitations become particularly pronounced in cluttered indoor environments where systematic coverage is essential.

Information-theoretic approaches employ metrics like Shannon entropy \cite{asgharivaskasi2023semantic} and mutual information \cite{stachniss2005information} to quantify and reduce mapping uncertainty through probabilistic modeling. Though theoretically sound, these methods face practical challenges including prohibitive computational costs for real-time information gain calculation, sensitivity to sensor noise, and frequent sacrifice of path efficiency for uncertainty reduction.

These traditional approaches establish important foundations but reveal significant gaps in handling the multi-objective nature of exploration. Their limitations in path optimality, computational efficiency, and adaptive decision-making motivate our development of a hybrid framework that preserves the strengths of systematic exploration while addressing these fundamental challenges through modern learning techniques. 
\subsection{DRL-based Active SLAM}
In the field of DRL-based robotic autonomous exploration, the design of reward functions and termination conditions constitutes the core decision-making mechanism of Active SLAM systems, directly influencing exploration efficiency and map quality. The following sections systematically review the research progress in these two critical aspects.
\subsubsection{Exploration Strategy Reward Function}
DRL-based exploration method has demonstrated significant potential in robotic autonomous exploration tasks, with the design of efficient reward functions being a core challenge to balance exploration efficiency and system robustness. Existing reward mechanisms can be categorized into three primary types: 
\paragraph{Map-Completeness-Based Reward Mechanisms} These methods motivate robots to achieve comprehensive environmental traversal through coverage increment incentives. For instance, Zhao et al.\cite{zhao2024exploration} decomposed rewards into map completeness, exploration rewards, and exploitation rewards to holistically incentivize exploration behaviors. Chaplot et al.\cite{chaplot2020learning} directly designed reward functions based on increases in covered area. These methods offer intuitive interpretability and ensure systematic environmental traversal. However, they often lead to suboptimal path planning in complex environments, particularly in obstacle-dense regions where robots may become trapped in inefficient repetitive exploration due to excessive focus on local coverage. 
\paragraph{Environment-Uncertainty-Based Reward Mechanisms} These methods leverage information entropy reduction or feature metrics of SLAM covariance matrices to drive active exploration. Chen et al.\cite{chen2024lidar} proposed a reward function integrating map information gain, control rewards, exploration completion rewards, and collision penalties. Alcalde et al.\cite{alcalde2022slam} and A. Placed et al.\cite{placed2020deep} adopted the D-optimality criterion to quantify localization and mapping uncertainties, embedding this metric into reward design. These approaches possess solid theoretical foundations and effectively reduce map uncertainty while improving localization accuracy. However, sensor noise can degrade performance, and path efficiency is often sacrificed to reduce uncertainty, resulting in excessive detours during exploration. 
\paragraph{Other Reward Mechanisms} Beyond the two primary mechanisms, several innovative reward designs have been proposed. For instance, Cao et al.\cite{cao2023ariadne} designed a composite reward function incorporating frontier point counts, path length penalties, and task completion incentives. In a related approach, Botteghi et al.\cite{botteghi2021curiosity} introduced an intrinsic curiosity-driven mechanism to encourage exploratory behavior. Similarly, Zhu et al.\cite{zhu2018deep} employed negative penalties proportional to path length to promote shorter, more efficient trajectories.

Notably, path length, as a critical metric of exploration efficiency, has been rarely systematically incorporated into reward function design in existing literature. While Cao et al.\cite{cao2023ariadne} and Zhu et al.\cite{zhu2018deep} introduced path length penalty terms in their respective works, no prior studies have proposed jointly optimizing path length and pose uncertainty as a dual-objective framework. The path-uncertainty co-optimization framework proposed in this study integrates both metrics into a unified reward function, dynamically balancing the trade-off between exploration and exploitation. This approach effectively addresses the suboptimal path planning issues resulting from existing methods' excessive focus on single objectives such as coverage area or map entropy, thereby providing a more comprehensive and effective decision-making mechanism for autonomous robotic exploration.
\subsubsection{Task Termination Mechanism}
The design of exploration task termination conditions is critical for ensuring both the completeness and computational efficiency of the exploration process. Existing methods primarily employ three termination mechanisms:
\paragraph{Environment-Triggered Termination Mechanisms}
Collision detection represents a typical example of this category, where tasks are terminated when the robot-obstacle distance falls below a predefined threshold (e.g., 0.2 meters) \cite{zhao2024exploration,chen2024lidar,botteghi2021curiosity,alcalde2022slam,placed2020deep}. This approach effectively prevents robots from continuing operation in hazardous environments, ensuring system safety. However, this mechanism lacks dynamic awareness of exploration progress, making it difficult to adapt to environments of varying complexity.
\paragraph{Task-Driven Termination Mechanisms}
Exploration completion thresholds (e.g., coverage ratio $\geq93\%$) have been validated and applied across multiple studies\cite{zhao2024exploration,cao2023ariadne,chen2024lidar,botteghi2021curiosity,placed2020deep,zhu2018deep,chaplot2020learning}. This mechanism ensures exploration tasks reach predefined objectives, but the threshold settings lack adaptability, making them unsuitable for environments of different scales and complexities. In simple environments, termination may occur too early; in complex environments, the threshold may never be reached, potentially resulting in indefinite task duration. Furthermore, this mechanism lacks dynamic awareness of exploration progress, making it difficult to adapt to environments of varying complexity.
\paragraph{Resource-Constrained Termination Mechanisms}
Fixed step limits or time ceilings ensure computational efficiency, as implemented in\cite{chen2024lidar,alcalde2022slam,chaplot2020learning}. These methods prevent indefinite exploration through predefined resource constraints but lack dynamic awareness of exploration progress. This can result in critical regions remaining unexplored before resource exhaustion or premature termination when resources are still available. 

Although some termination conditions have been designed in the above-mentioned papers, no systematic approach has been developed to detect and correct abnormal exploration behaviors that lead to inefficient stagnation. The absence of such mechanisms can cause significant performance degradation, particularly in complex environments where robots may become trapped in local oscillations or inefficient wandering patterns. The lightweight stagnation detection mechanism proposed in this paper, through real-time LiDAR analysis and adaptive map-update monitoring, effectively identifies and corrects these problematic behaviors. This approach not only enhances the robustness of the exploration process but also provides a crucial missing component for comprehensive termination condition design in Active SLAM systems.
\section{APPROACH}
\begin{figure}
	\centering         
	\includegraphics[width=\textwidth]{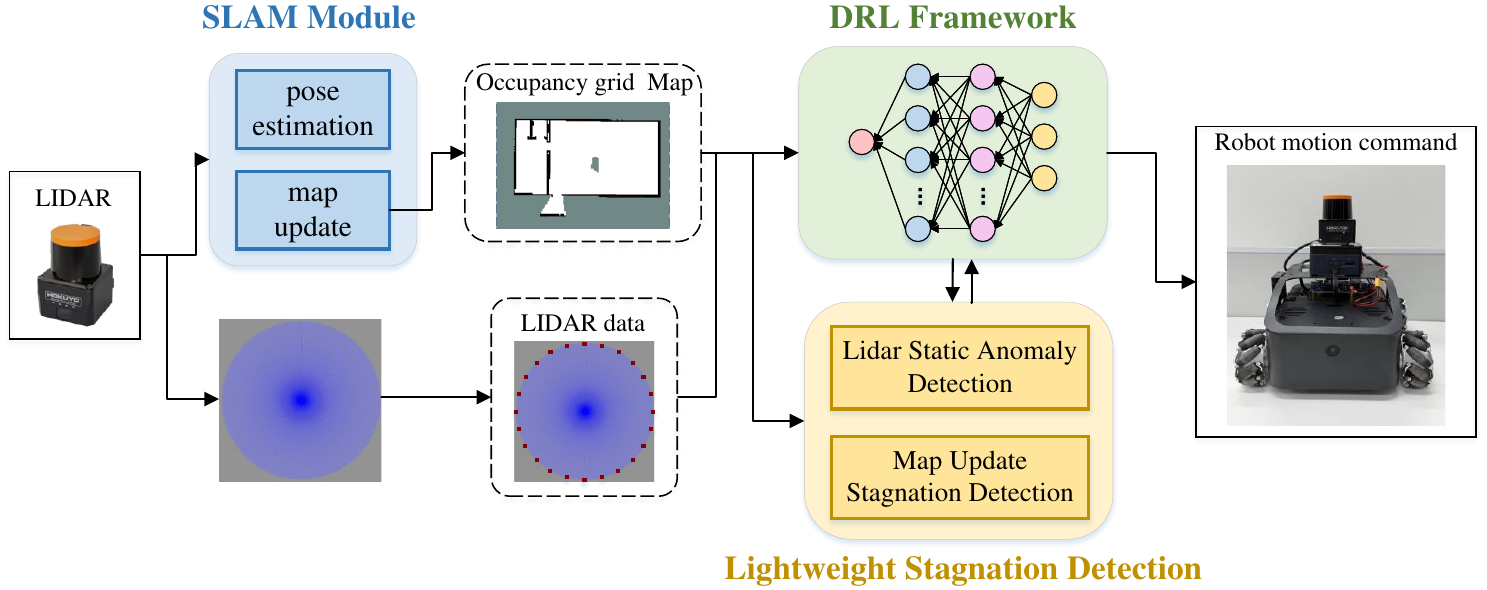}     
	\caption{The overall framework of the PUL-SLAM system.}
	\label{fig:frame}
\end{figure}

As illustrated in Fig. \ref{fig:frame}, the PUL-SLAM system features a dual-layer collaborative optimization architecture, integrating a high-level DRL-based decision-making framework with a low-level Lightweight Stagnation Detection mechanism. In the operational pipeline, the SLAM module first processes LiDAR inputs to generate a real-time occupancy grid map alongside robot pose estimation. Subsequently, the Lightweight Stagnation Detection module, which consists of LiDAR Static Anomaly Detection and Map Update Stagnation Detection, monitors for inefficient behaviors. The former identifies motion anomalies via consecutive frame similarity analysis, while the latter tracks map expansion rates. If any abnormal state is identified, the system promptly halts the current exploration episode and performs an environment reset to avoid reinforcing inefficient strategies. Simultaneously, the DRL framework leverages a path-uncertainty co-optimization strategy to dynamically trade off exploration efficiency with pose uncertainty, thereby producing optimal motion commands to drive the robot.
\subsection{Path-Uncertainty Co-Optimization DRL Framework}
\subsubsection{Reward Function}
The reward function, as the core mechanism guiding agent learning in reinforcement learning, directly determines the performance of the algorithm. The proposed path-uncertainty co-optimization reward function aims to achieve a dynamic balance between exploration and exploitation through a dual-objective optimization mechanism, thereby avoiding the inefficiency problems caused by excessive focus on a single objective in traditional methods. Specifically, the reward function is formulated as:
\begin{equation}
	\mathcal{R}_{t}=\begin{cases}
		1+\text{tanh\ensuremath{\left(\frac{\eta}{f\left(\Sigma\right)}\right)}+\ensuremath{\mathcal{P}_{t}}} & \text{if }\Delta c_{t}>0\\
		0.001+\mathcal{P}_{t} & \text{else if }\neg\text{done}\\
		-100 & \text{otherwise}
	\end{cases},\label{eq:1}
\end{equation}
where $\eta$ is a task-dependent scale factor, $f\left(\Sigma\right)$ is the D-optimality criterion\cite{carrillo2012comparison,alcalde2022slam}, $\Delta c_{t}$ represents the newly added map area at time $t$, and $\mathcal{P}_{t}$ is the path penalty term, which is defined as follows:
\begin{equation}
	\mathcal{P}_{t}=\begin{cases}
		-0.1*d_{t} & \text{if }\eta_{t}<0.001\text{ and }d_{t}>0.001\\
		0 & \text{otherwise}
	\end{cases},\label{eq:2}
\end{equation}
where the exploration efficiency $\eta_{t}$ is defined as the ratio of the newly added map area $\Delta c_{t}$ to the robot's incremental distance $d_{t}$, where $d_{t}$ denotes the distance traversed by the robot from time $t - 1$ to $t$. A path penalty is imposed only when $\eta_{t}$ falls below a predefined threshold.

\subsubsection{Observation Space}
\begin{figure}
	\centering         
	\includegraphics[width=0.8\textwidth]{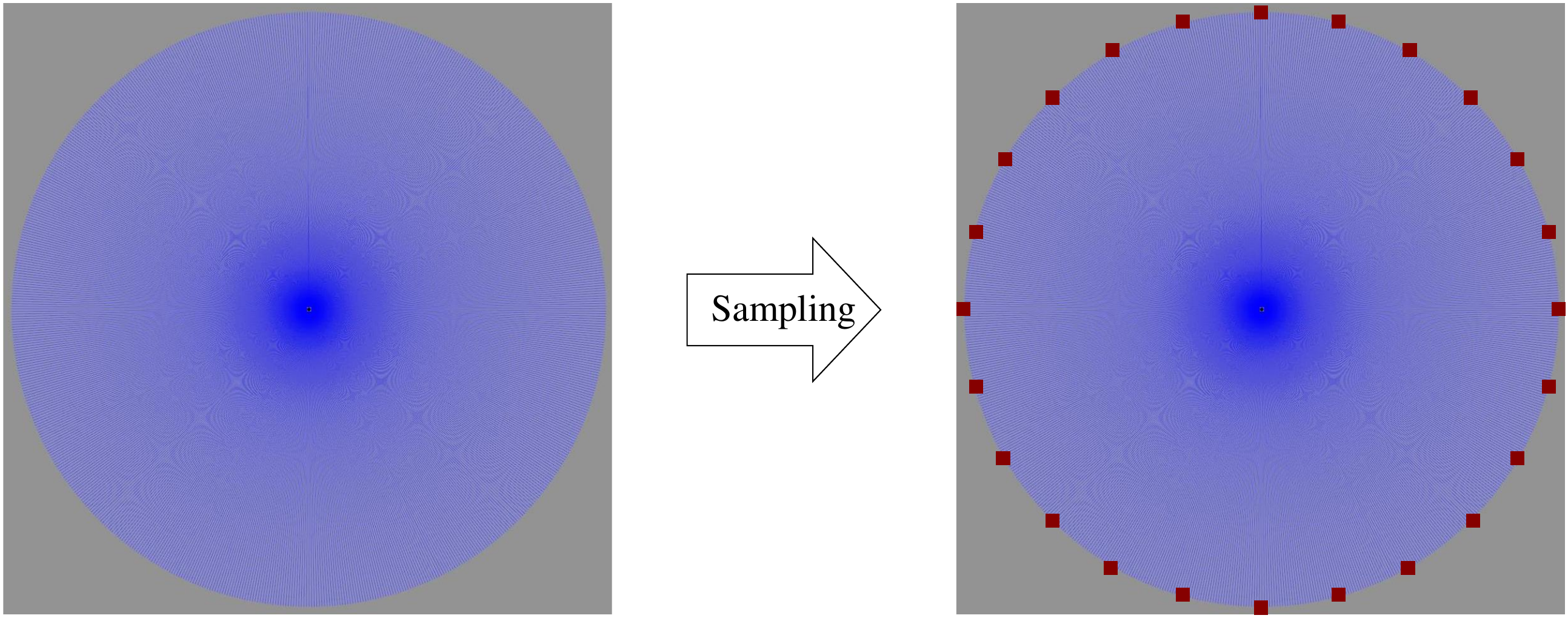}     
	\caption{For the sampling of LiDAR data, the data volume changes from 360 to $N$.}
	\label{fig:lidar}
\end{figure}

At time step $t$, we uniformly sample $360$ laser measurements to obtain $N$ ranging values, as illustrated in Fig. \ref{fig:lidar}, yielding range values normalized to: 	\begin{equation}
	\hat{\textbf{s}}_{t}=\left[\hat{d}_{t}^{\left(1\right)},\hat{d}_{t}^{\left(2\right)},\dots,\hat{d}_{t}^{\left(N\right)}\right]^{\intercal}\in\left[0,1\right]^{N}.  
\end{equation}

By reducing the number of sampling points $N$, computational complexity is effectively reduced while maintaining sufficient environmental representational capacity, allowing the algorithm to run in real-time on resource-constrained mobile robot platforms.

The observation space $\textbf{O}_{t}$ for the reinforcement learning agent consists of two components: the normalized laser scan vector $\hat{\textbf{s}}_{t}$ and the current map coverage ratio $c_{t}$:
\begin{equation}
	{\textbf{O}}_{t}=\left[\hat{\textbf{s}}_{t},c_{t}\right].  
\end{equation}

This design choice integrates local perception with global state awareness: the laser scan vector captures fine-grained geometric details of the immediate surroundings, furnishing the agent with real-time sensory input for decision-making, while the cumulative map coverage ratio serves as a global indicator of exploration progress, thereby enabling the agent to maintain a coherent understanding of its spatial context.
\subsubsection{Action Space}
Based on the discrete action space strategy, the robot's kinematic control parameters for three fundamental motion commands are defined as follows:
\begin{itemize}
	\item Forward: Linear velocity is set at $v = 0.2  \text{m/s}$ with angular velocity $\omega = 0  \text{rad/s}$, ensuring linear motion along the current heading direction.
	\item Turn left: Linear velocity $v = 0.2  \text{m/s}$ and angular velocity $\omega = 0.4  \text{rad/s}$, generating a smooth left-turning trajectory.
	\item Turn right: Linear velocity $v = 0.2  \text{m/s}$ and angular velocity $\omega = -0.4  \text{rad/s}$, producing a symmetric right-turning behavior.
\end{itemize}
\subsubsection{Neural Networks}
We adopt the Proximal Policy Optimization (PPO)\cite{schulman2017proximal} algorithm for policy learning. The observation space is structured as a dictionary comprising $N$-dimensional normalized LiDAR readings and a 1-dimensional map coverage metric, resulting in a ($N+1$)-dimensional joint input vector. This concatenated observation is processed by a shared backbone network consisting of two fully connected layers, each with 64 neurons and Tanh activation functions. The shared representation is then fed into two separate heads: a policy head that outputs logits over three discrete actions (forward, turn left, turn right), and a value head that estimates the scalar state-value function. All network parameters are jointly optimized in an end-to-end manner via gradient-based updates.
\subsection{Lightweight Stagnation Detection Module}
\subsubsection{Lidar Static Anomaly Detection}
Robotic exploration can exhibit intentional pausing (e.g., stationary behavior to maximize reward in RL) or motion failure (e.g., collisions or wheel slippage). Undetected motion failures degrade learning efficiency by slowing convergence and reinforcing suboptimal policies. To address this, we propose Lidar Static Anomaly Detection: a lightweight method that identifies motion failures via cosine similarity between consecutive LiDAR scans. Unlike odometry/IMU-based approaches, this method uses raw environmental perception data, maintaining reliability during motor idling or wheel slippage. By operating on normalized scan vectors, it inherently rejects localized environmental changes. Lidar Static Anomaly Detection achieves real-time stagnation detection, with negligible impact on exploration performance.

The similarity metric between two consecutive frames of processed LiDAR data at timestamps $t$ and $t-1$ is calculated as follows:
\begin{equation}
	\cos\left(t\right)=\frac{\langle\hat{\textbf{s}}_{t},\hat{\textbf{s}}_{t-1}\rangle}{\lVert\hat{\textbf{s}}_{t}\rVert\cdot\lVert\hat{\textbf{s}}_{t-1}\rVert},\label{eq:6}
\end{equation}
where $\langle\cdot,\cdot\rangle$ denotes the vector inner product operation, and $\lVert\cdot \lVert$ represents the Euclidean norm. Consequently, the above expression expands to:
\begin{equation}
	\cos\left(t\right)=\frac{\sum_{i=1}^{N}\hat{d}_{t}^{\left(i\right)}\hat{d}_{t-1}^{\left(i\right)}}{\sqrt{\sum_{i=1}^{N}\left(\hat{d}_{t}^{\left(i\right)}\right)^{2}}\sqrt{\sum_{i=1}^{N}\left(\hat{d}_{t-1}^{\left(i\right)}\right)^{2}}}.\label{eq:7}
\end{equation}

A static indicator function is formally defined as:
\begin{equation}
	\mathbb{I}\left(t\right)=\begin{cases}
		1 & \text{if }\cos\left(t\right)>\alpha\\
		0 & \text{otherwise}
	\end{cases},\label{eq:8}
\end{equation}
where $\alpha\in\left(0,1\right)$ is the similarity threshold. Consequently, the static state counter $C$ updates according to the following rule:
\begin{equation}
	\mathcal{C}\left(t\right)=\begin{cases}
		\mathcal{C}\left(t-1\right)+1 & \text{if }\mathbb{I}\left(t\right)=1\\
		0 & \text{otherwise}
	\end{cases}.\label{eq:9}
\end{equation}

The static status flag static $\mathcal{F}$ is a Boolean signal derived from the state counter, formalized as:
\begin{equation}
	\mathcal{F}\left(t\right)=\begin{cases}
		1 & \text{if }\mathcal{C}\left(t\right)\geq\Omega\\
		0 & \text{otherwise}
	\end{cases},\label{eq:10}
\end{equation}
where $\Omega\in\mathbb{Z}^{+}$ is the continuity threshold.
\subsubsection{Map Update Stagnation Detection}
In autonomous exploration tasks, robots may fall into ineffective exploration states due to various reasons, including but not limited to wheel slippage, sensor malfunctions, or lack of distinctive environmental features. These stagnation states not only significantly reduce exploration efficiency but may also lead to reward hacking, where the algorithm learns to maximize cumulative rewards by remaining stationary for extended periods rather than conducting genuine exploration. To address this issue, we propose an innovative Map Update Stagnation Detection mechanism that identifies and terminates ineffective exploration behaviors through real-time monitoring of map expansion rate while ensuring legitimate low-speed exploration in sparse environments remains undisturbed.

The core concept of map update stagnation detection is to quantify the increment of newly explored area per unit time as an objective metric of exploration efficiency. Let $\Delta t$ denote the corresponding time interval. The map expansion rate can then be defined as:
\begin{equation}
	\dot{c}_t = \Delta c_t / \Delta t,
\end{equation}

This metric directly reflects the robot's exploration efficiency: when the robot is in an effective exploration state, $\dot{c}_t$ should remain within a certain positive range; when the robot enters a stagnation state, $\dot{c}_t$ will approach zero.

However, simply setting a fixed threshold cannot distinguish between genuine motion failures and legitimate low-speed exploration in sparse environments. To address this limitation, we designed a dynamic detection mechanism with time-cumulative effects. Let $\epsilon$ represent the environment-adaptive minimum effective exploration rate threshold, and T denote the continuous stagnation detection time window. The stagnation state can then be formally defined as:
\begin{equation}
	S(t) = 
	\begin{cases} 
		1 & \text{if } (\forall \tau \in [t-T, t], \hat{c}_\tau < \epsilon) \land (\|\mathbf{v}\| > 0) \\
		0 & \text{otherwise}
	\end{cases},
\end{equation}
where $\|\mathbf{v}\|$ denotes the Euclidean norm of the velocity vector, representing the robot's actual movement magnitude. When $S(t)=1$, the system determines that the robot has entered a stagnation state and triggers appropriate corrective measures. Notably, we specifically included the condition that velocity commands are active to ensure that stagnation detection only occurs when the robot is actively attempting to move, preventing legitimate boundary point pauses from being misclassified as stagnation.

For the design of threshold $\epsilon$, we propose an environment-characteristic-based adaptive calculation method:
\begin{equation}
	\epsilon = \beta \cdot \frac{A_{\text{env}}}{T_{\text{max}}},
\end{equation}
where $A_{\text{env}}$ represents the estimated environment area, $T_{\text{max}}$ denotes the typical exploration time for the environment, and $\beta$ is an empirical coefficient. This design ensures the threshold can adapt to environments of varying scales and complexities: in expansive environments, the threshold is higher to prevent premature termination of legitimate exploration; in narrow environments, the threshold is lower, enabling the system to more sensitively detect motion failures.

\section{EXPERIMENTS AND RESULTS}
\subsection{Experimental Settings}
\begin{figure}
	\centering         
	\includegraphics[width=0.8\textwidth]{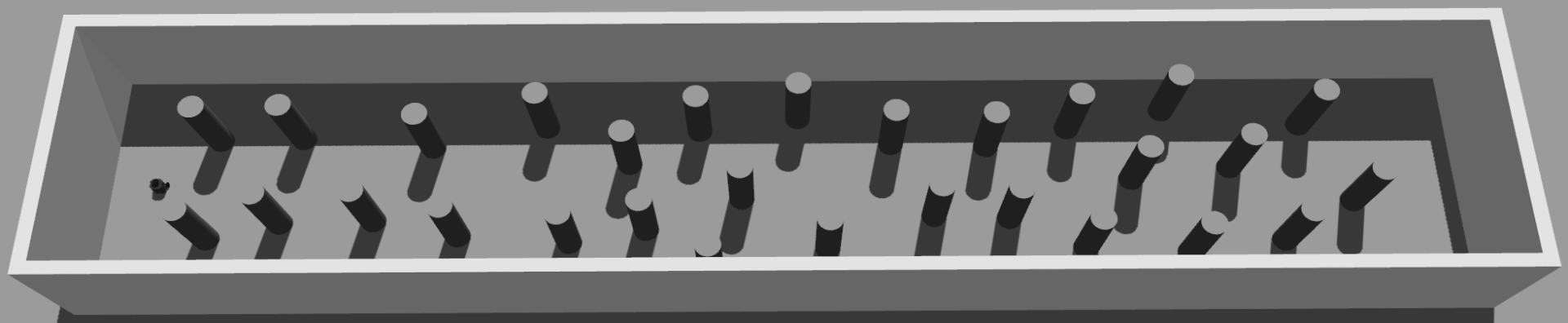}     
	\caption{Env-1 for training.}
	\label{fig:Env1}
\end{figure}

To validate our algorithm, we established a ROS\cite{quigley2009ros}-based simulation platform on an Ubuntu 20.04 system. This platform utilizes the Gazebo simulator to replicate realistic physical scenarios. Experiments were conducted using the TurtleBot3-Burger robot, equipped with a 360-degree LiDAR (maximum ranging distance of 3.5 meters) and wheel odometry for motion tracking. As shown in Fig. \ref{fig:Env1}, the training scenario is a rectangular room with dimensions of 15 meters by 3 meters, filled with cylindrical obstacles. The density of these obstacles increases gradually from left to right, creating a gradient of complexity within the environment. The robot model initiates its learning and exploration process from the left side of the room. In the testing phase, three additional testing scenarios were introduced. These testing scenarios cover a range of areas from 56 to 128 square meters, representing environments of varying scales and complexities. Among them, Env-2 and Env-3 are standard scenarios commonly used for validating Active SLAM algorithms, and some studies\cite{alcalde2022slam,placed2020deep} have already conducted experiments in these environments. Env-4 is a complex suite designed by us, containing multiple obstacles with a more intricate layout. It is intended to further verify the robot's adaptability and robustness in unknown and complex environments. By conducting tests in these diverse scenarios, we can comprehensively evaluate the performance of the proposed decision-making algorithm under different environmental conditions. The main parameters used in the experiment are shown in Table \ref{param}.

\begin{table}[h]
	%\vspace{-0.2cm}
	\caption{Training and simulation main hyperparameters.}
	\centering
	\begin{tabular}{cccc}
		\toprule  % ?????
		Hyperparameters & value \\
		\midrule  % ????? 
		Batch size & 64 \\
		Max episode steps & 5000 \\
		Training iteration & 350000 \\
		Discount factor $\gamma$  & 0.99 \\
		Learning rate & 0.0003 \\
		Scale factor $\eta$ & 1 \\
		Number of LiDAR samples $N$ & 24 \\
		Similarity threshold $\alpha$ & 0.98 \\
		Continuity threshold $\Omega$ & 10 \\
		Time interval $T$ & 20 \\
		Stagnation threshold factor $\beta$ & 0.05 \\
		\bottomrule  % ?????
	\end{tabular}
	\label{param}
\end{table}

\begin{figure}[htbp]
	\centering
	\begin{subfigure}[b]{0.38\textwidth}
		\centering
		\includegraphics[width=\textwidth]{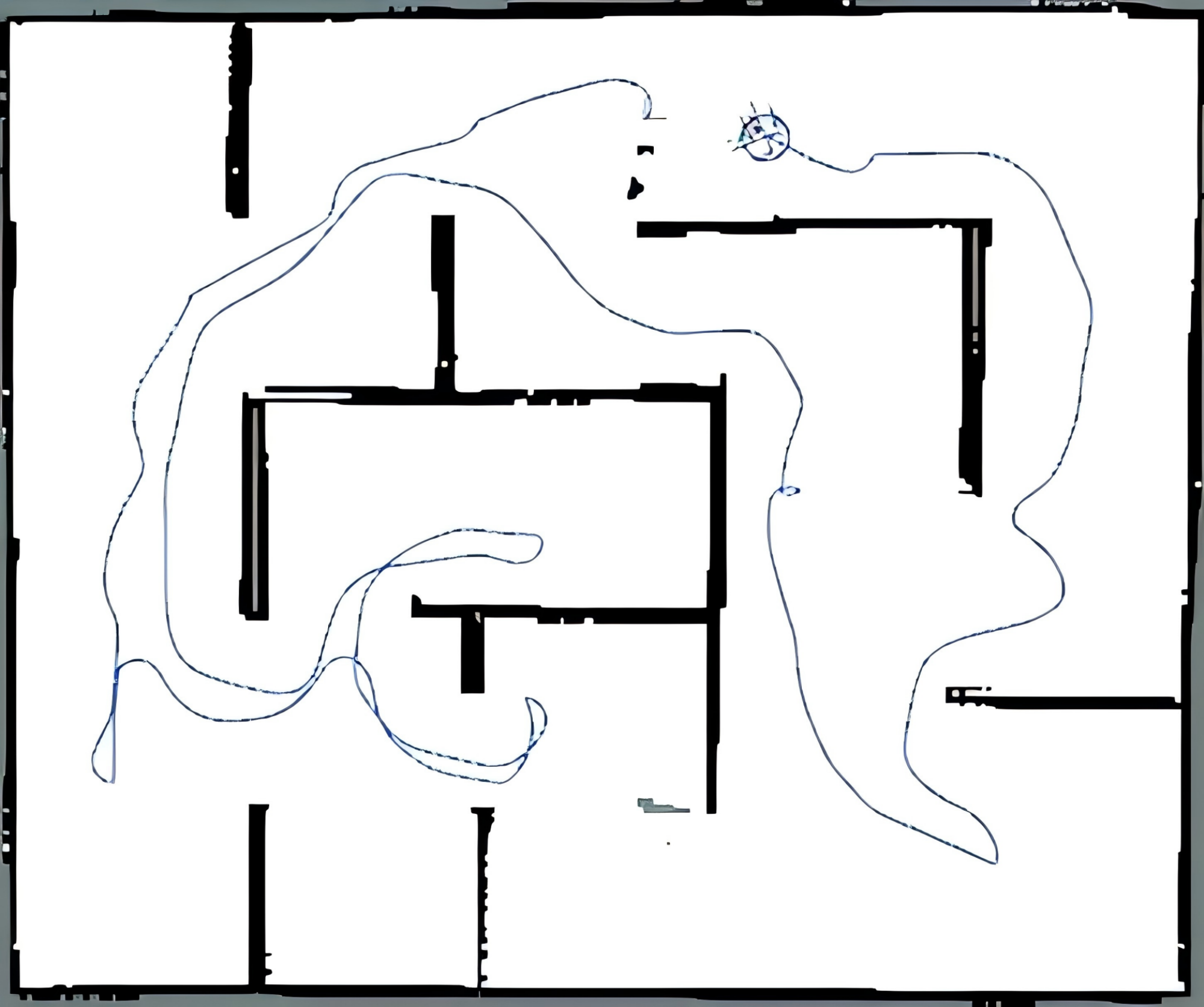}
		\caption{Frontier}
	\end{subfigure}%
	\hspace{2em}% 添加一个适中的固定空隙（可调）
	\begin{subfigure}[b]{0.38\textwidth}
		\centering
		\includegraphics[width=\textwidth]{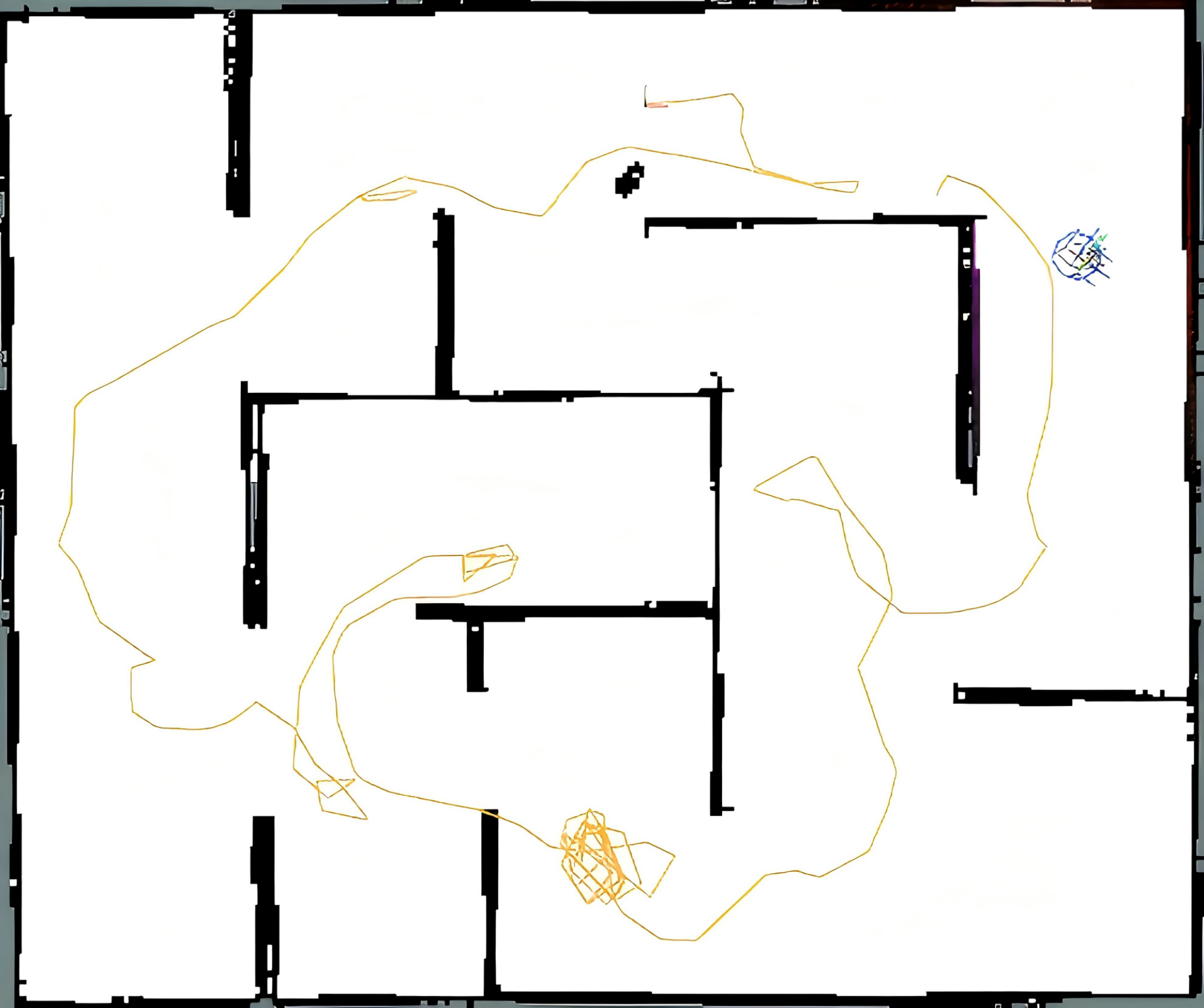}
		\caption{RRT}
	\end{subfigure}
	\vspace{-0.35cm}
	\vskip\baselineskip % 换行（模拟第二行）
	\begin{subfigure}[b]{0.37\textwidth}
		\centering
		\includegraphics[width=\textwidth]{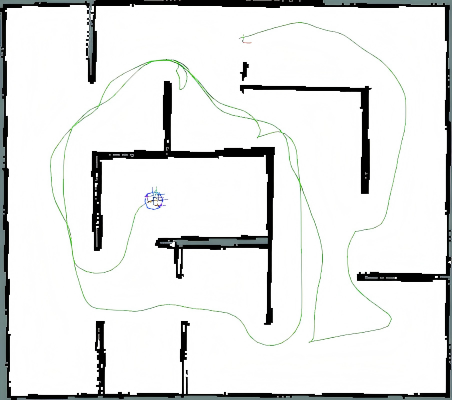}
		\caption{DA-SLAM}
	\end{subfigure}%
	\hspace{2em}% 同样添加空隙
	\begin{subfigure}[b]{0.39\textwidth}
		\centering
		\includegraphics[width=\textwidth]{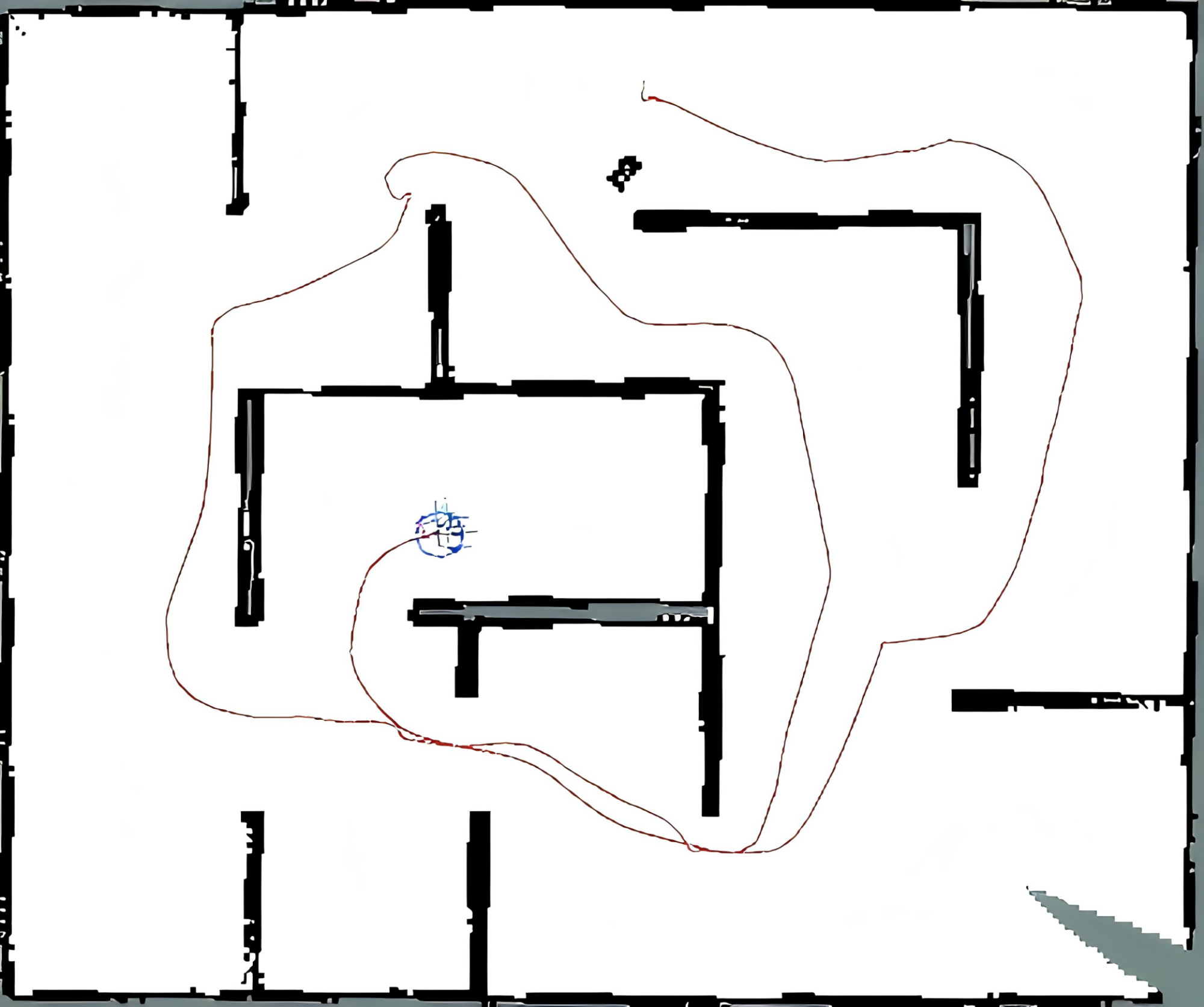}
		\caption{Ours}
	\end{subfigure}
	\caption{Trajectory and mapping results in Env-2.}
	\label{fig:env2}
\end{figure}

\begin{figure}[htbp]
	\centering
	\begin{subfigure}[b]{0.42\textwidth}
		\centering
		\includegraphics[width=\textwidth]{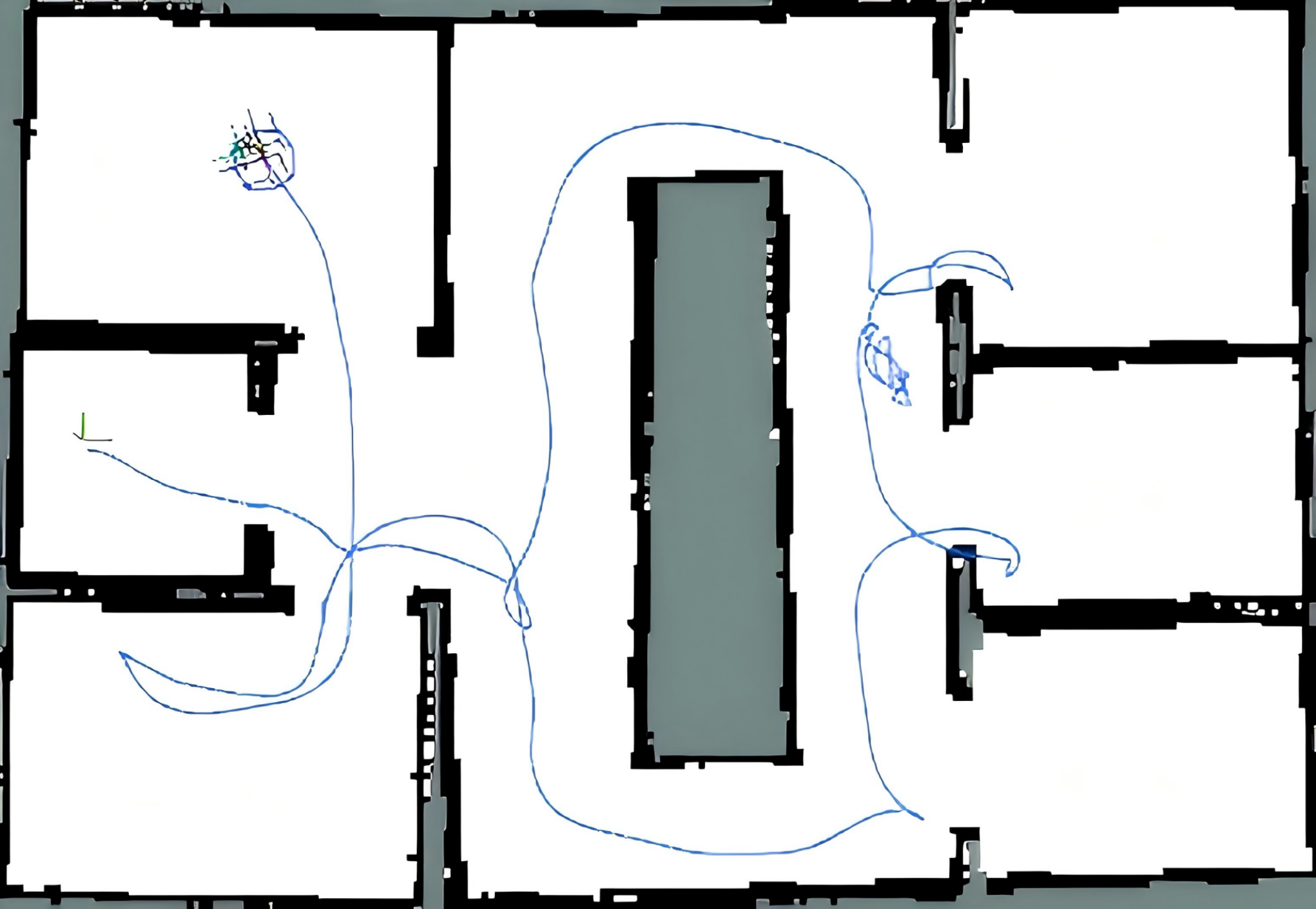}
		\caption{Frontier}
	\end{subfigure}%
	\hspace{2em}% 添加一个适中的固定空隙（可调）
	\begin{subfigure}[b]{0.42\textwidth}
		\centering
		\includegraphics[width=\textwidth]{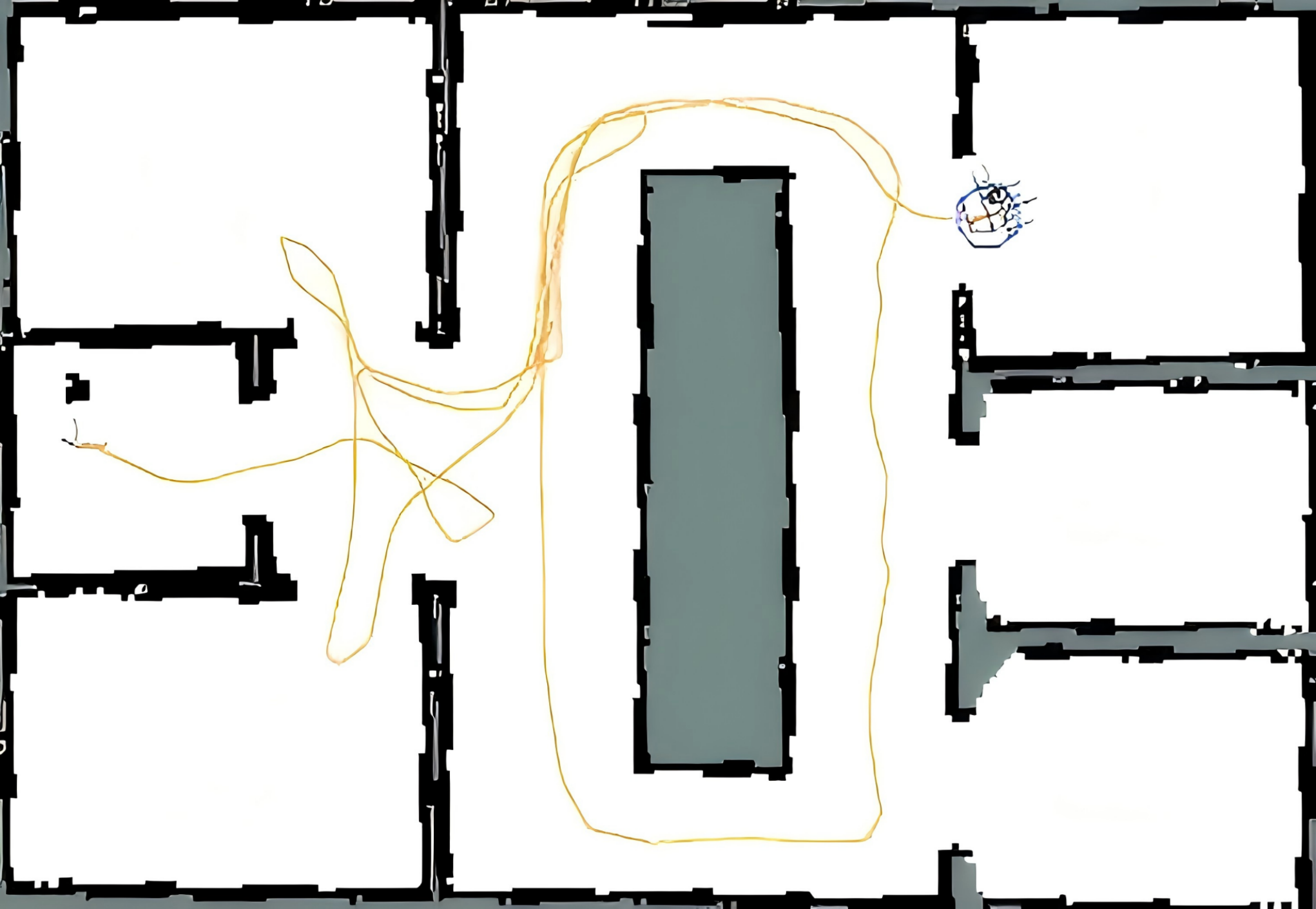}
		\caption{RRT}
	\end{subfigure}
	\vspace{-0.35cm}
	\vskip\baselineskip % 换行（模拟第二行）
	\begin{subfigure}[b]{0.42\textwidth}
		\centering
		\includegraphics[width=\textwidth]{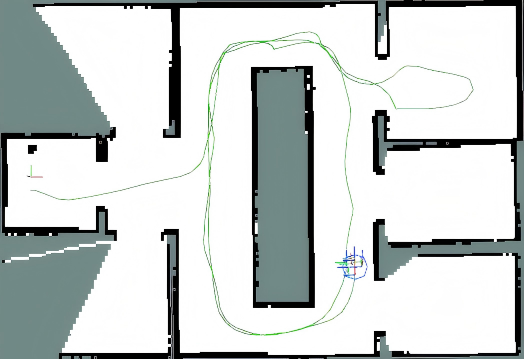}
		\caption{DA-SLAM}
	\end{subfigure}%
	\hspace{2em}% 同样添加空隙
	\begin{subfigure}[b]{0.42\textwidth}
		\centering
		\includegraphics[width=\textwidth]{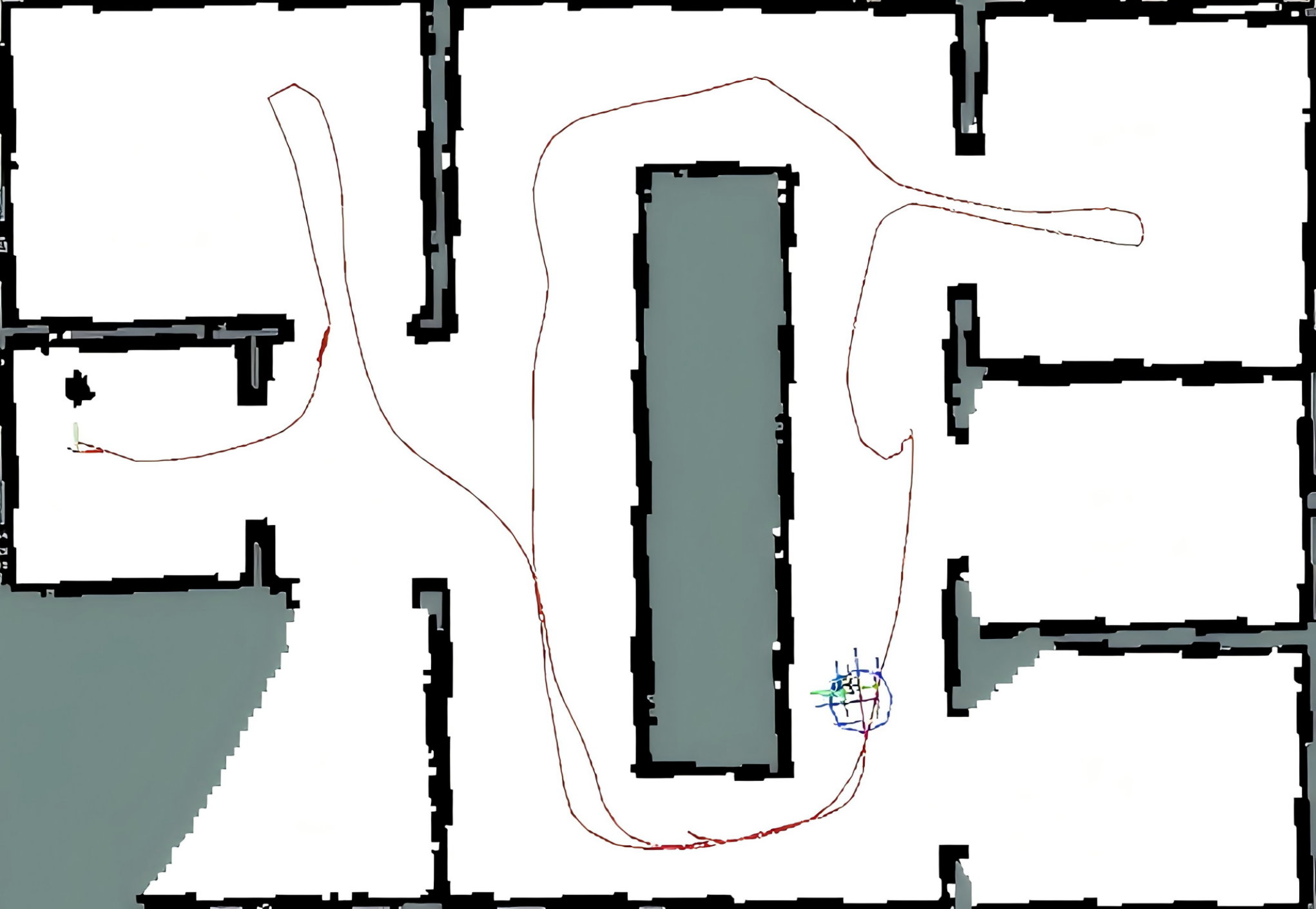}
		\caption{Ours}
	\end{subfigure}
	\caption{Trajectory and mapping results in Env-3.}
	\label{fig:env3}
\end{figure}

\begin{figure}[htbp]
	\centering
	\begin{subfigure}[b]{0.45\textwidth}
		\centering
		\includegraphics[width=\textwidth]{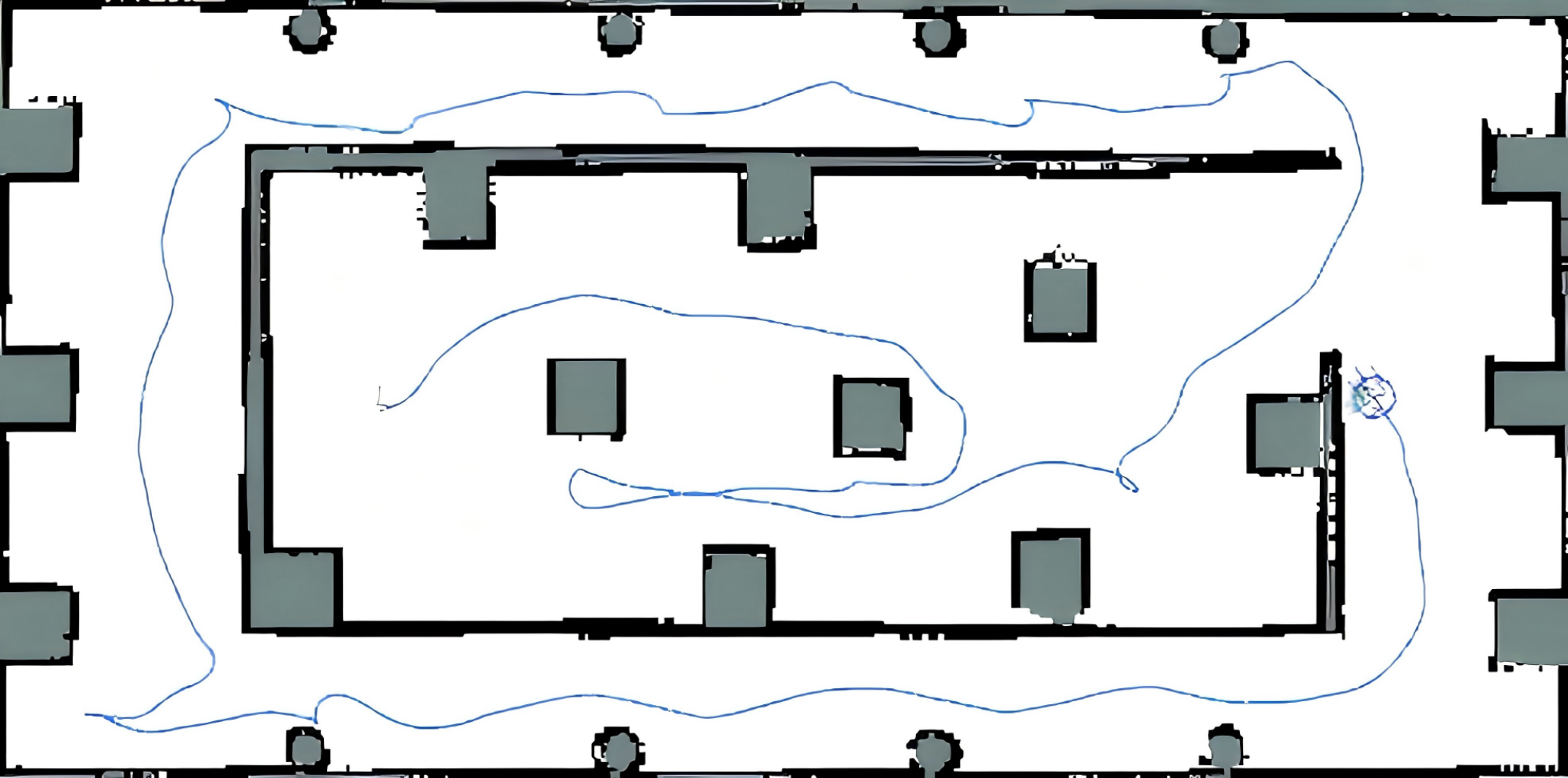}
		\caption{Frontier}
	\end{subfigure}%
	\hspace{2em}% 添加一个适中的固定空隙（可调）
	\begin{subfigure}[b]{0.45\textwidth}
		\centering
		\includegraphics[width=\textwidth]{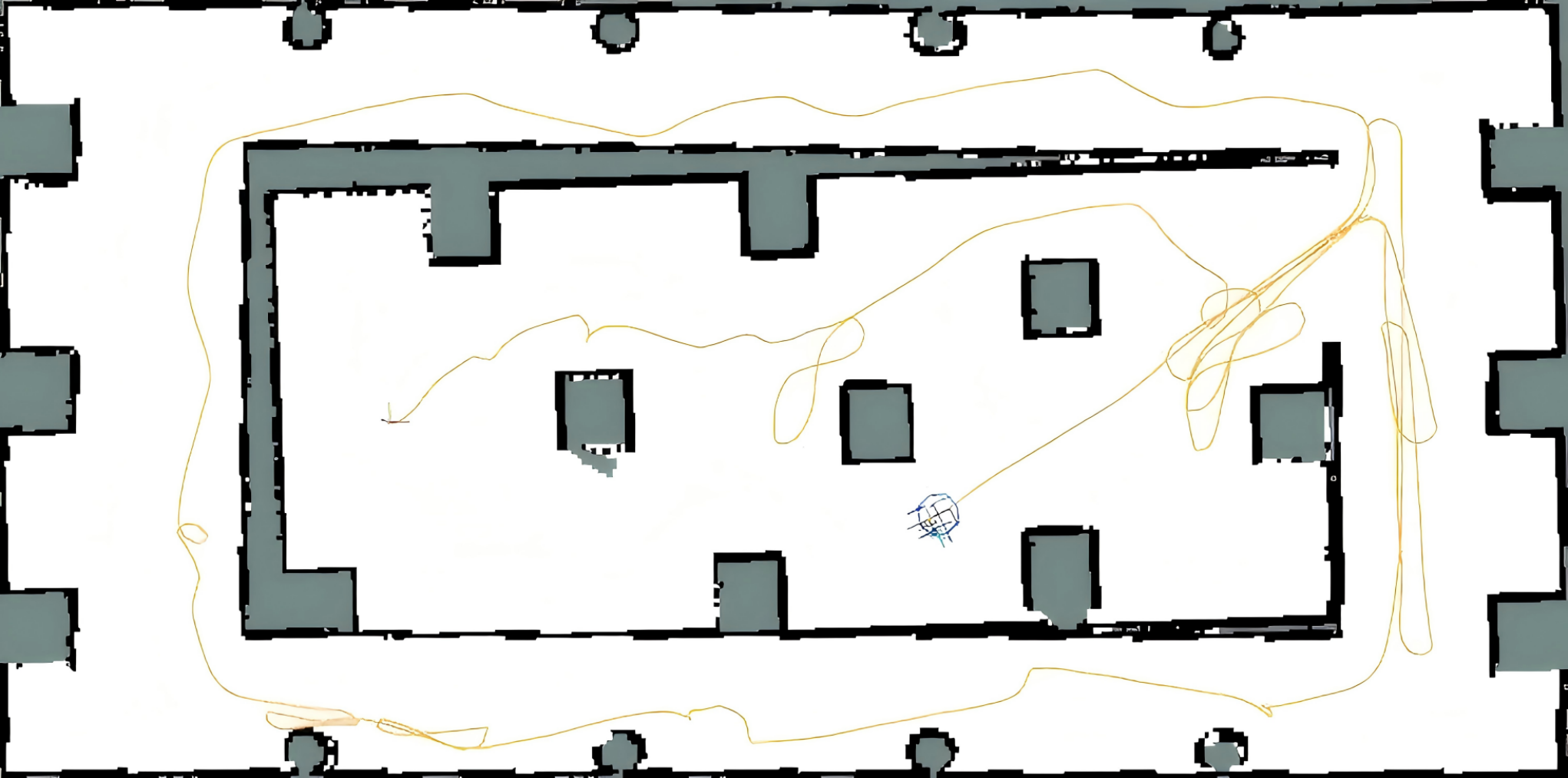}
		\caption{RRT}
	\end{subfigure}
	\vspace{-0.35cm}
	\vskip\baselineskip % 换行（模拟第二行）
	\begin{subfigure}[b]{0.45\textwidth}
		\centering
		\includegraphics[width=\textwidth]{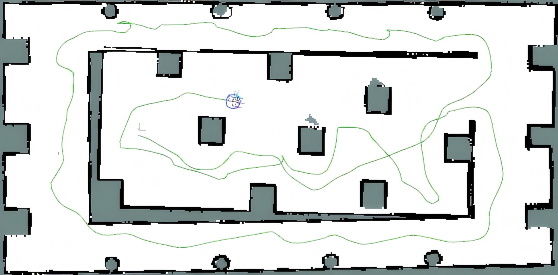}
		\caption{DA-SLAM}
	\end{subfigure}%
	\hspace{2em}% 同样添加空隙
	\begin{subfigure}[b]{0.45\textwidth}
		\centering
		\includegraphics[width=\textwidth]{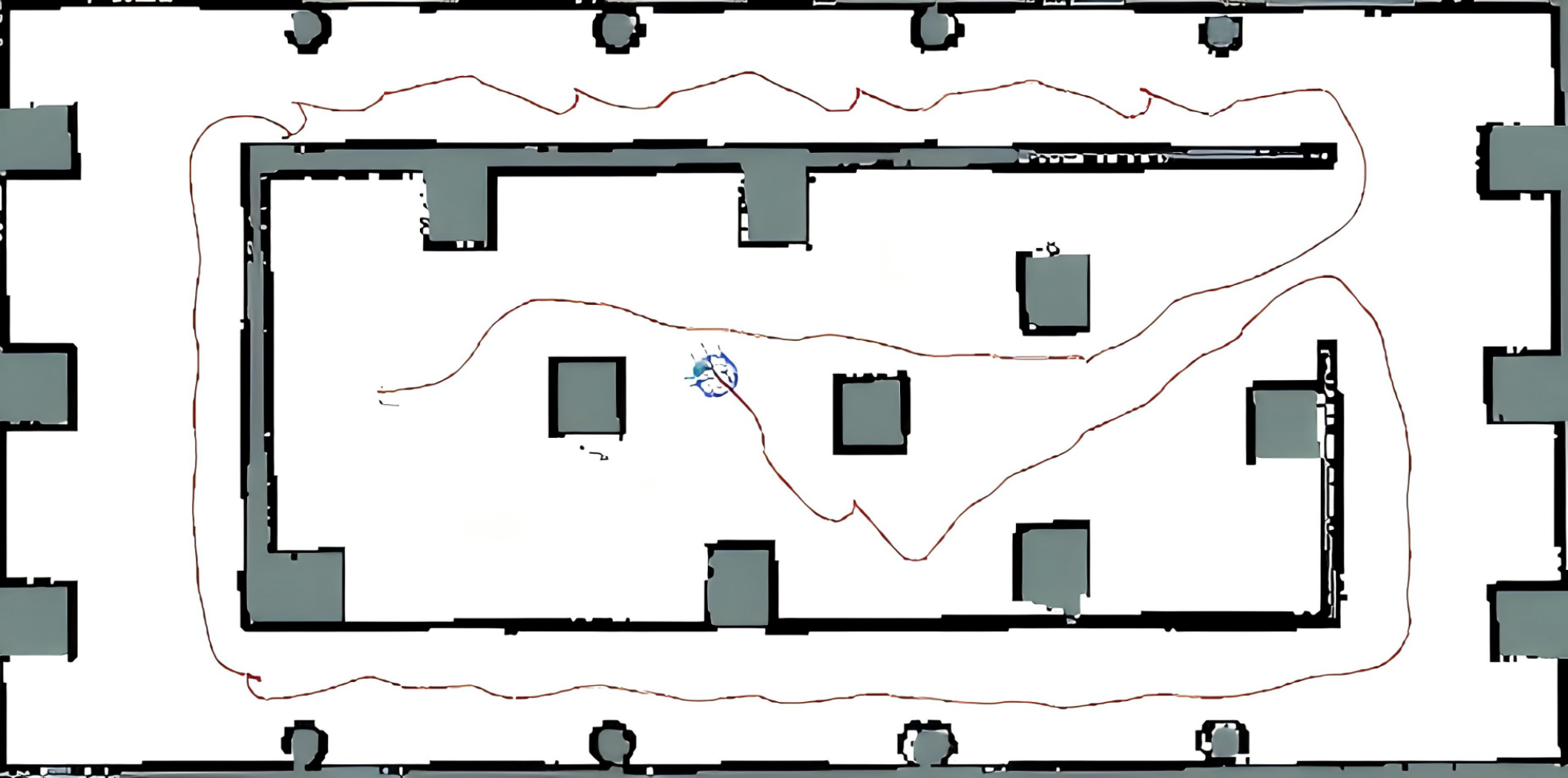}
		\caption{Ours}
	\end{subfigure}
	\caption{Trajectory and mapping results in Env-4.}
	\label{fig:env4}
\end{figure}

\begin{table}
	\centering
	\setlength{\abovecaptionskip}{0.1cm}
	\caption{Evaluation results in Env-2, Env-3, Env-4.}
	\renewcommand\arraystretch{0.9}
	\begin{tabularx}{\textwidth}{>{\centering\arraybackslash}X >{\centering\arraybackslash}X c >{\centering\arraybackslash}X >{\centering\arraybackslash}X >{\centering\arraybackslash}X}
		\toprule
		Env & \makecell{Scenario} & Method & \makecell{Time(s)} & \makecell{Path\\Length(m)} & \makecell{Map\\Completeness\\(\%)} \\ 
		
		\midrule
		\multirow{4}{*}{Env-2} & \multirow{4}{*}{\includegraphics[width=0.11\textwidth]{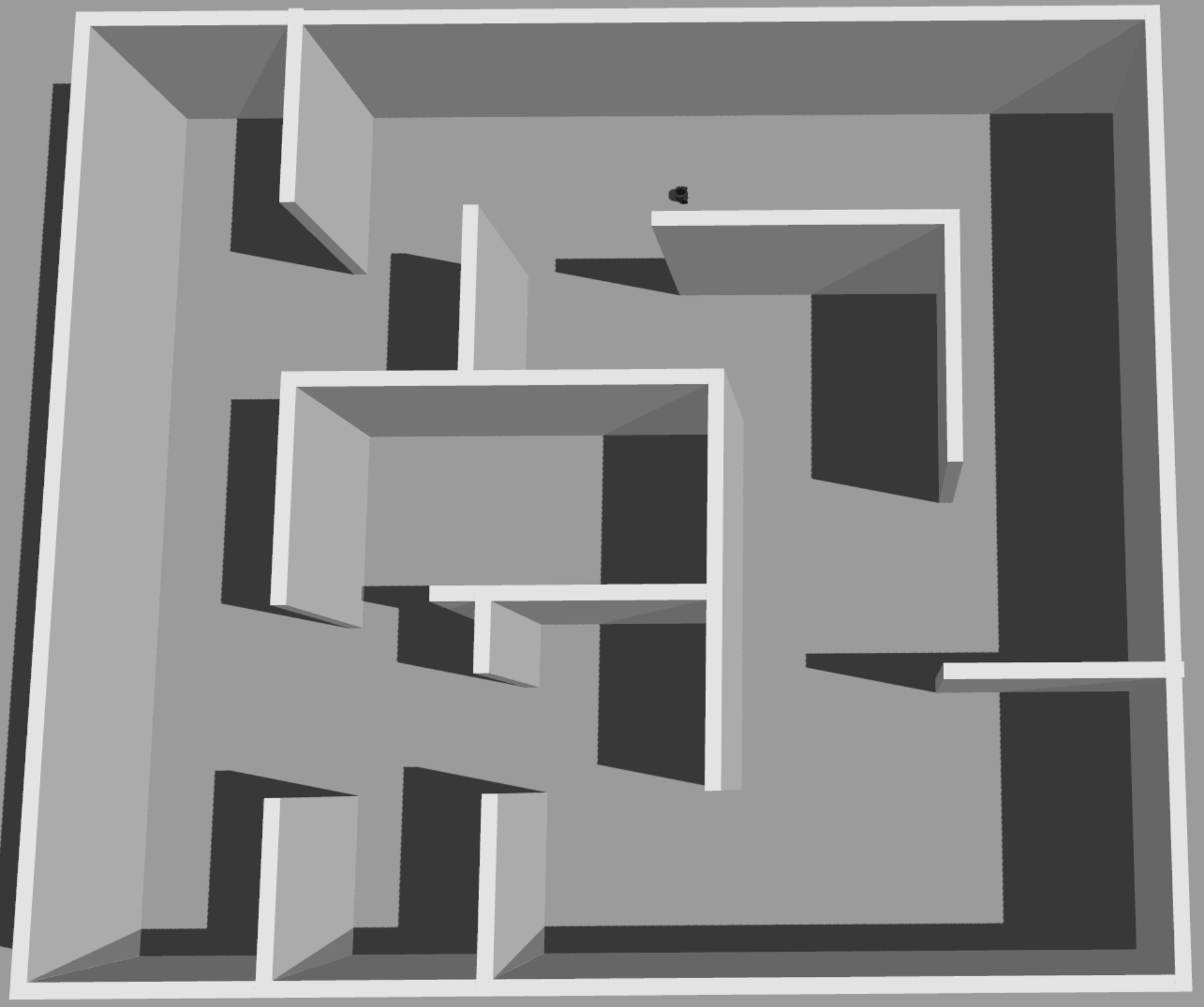}} & Frontier & 322.14 & 50.19 & 99.30\\
		& & RRT & 553.68 & 58.40 & 99.24\\
		& & DA-SLAM & 288.84 & 51.88 & \textbf{100}\\
		& & Ours & \textbf{235.23} & \textbf{41.84} & 98.56\\
		
		\midrule
		\multirow{4}{*}{Env-3} & \multirow{4}{*}{\includegraphics[width=0.13\textwidth]{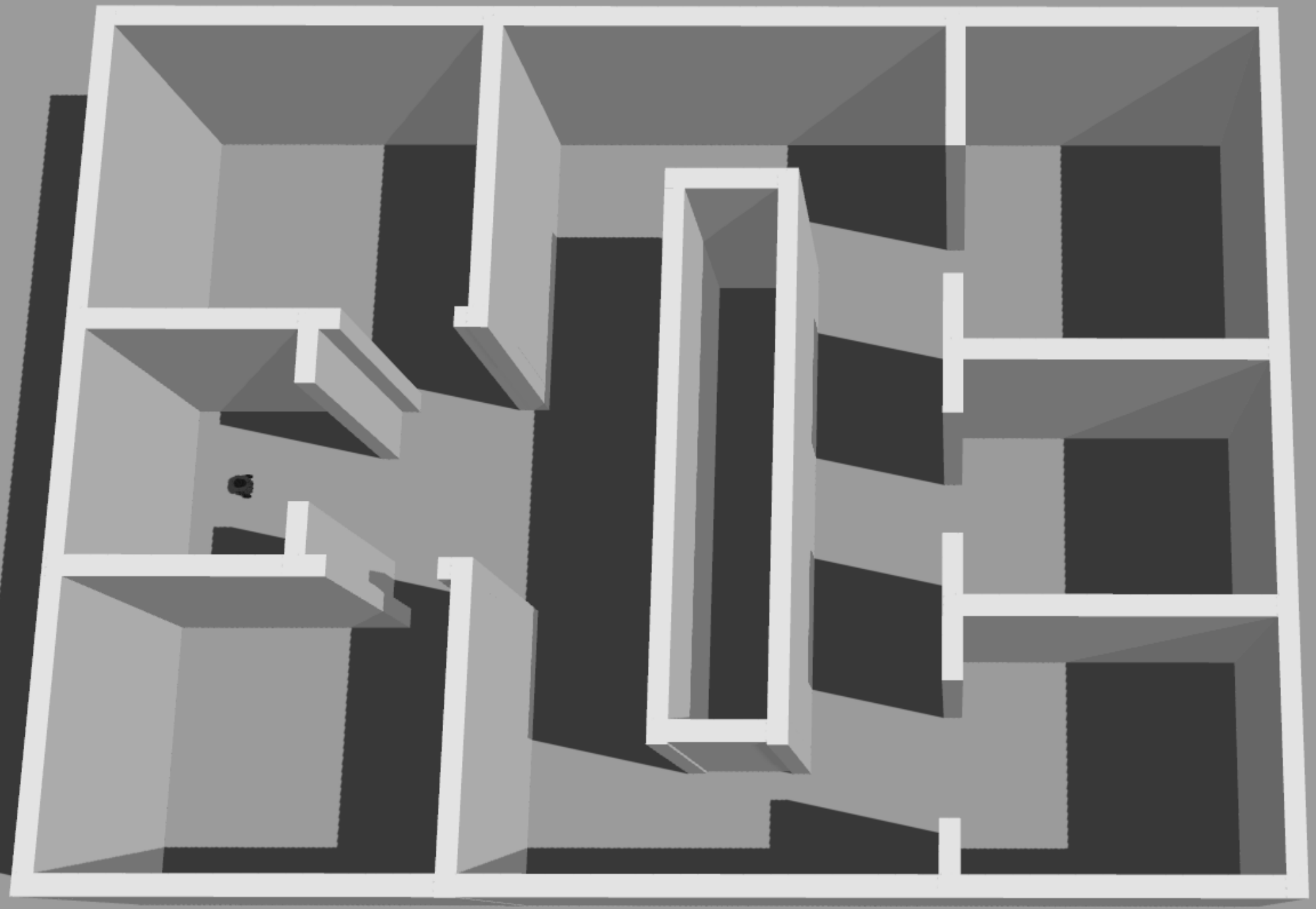}} & Frontier & 334.98 & 34.10 & \textbf{99.68}\\
		& & RRT & 268.89 & 37.29 & 98.45\\
		& & DA-SLAM & 199.59 & 36.12 & 89.23\\
		& & Ours & \textbf{167.07} & \textbf{29.26} & 94.13\\
		
		\midrule
		\multirow{4}{*}{Env-4} & \multirow{4}{*}{\includegraphics[width=0.15\textwidth]{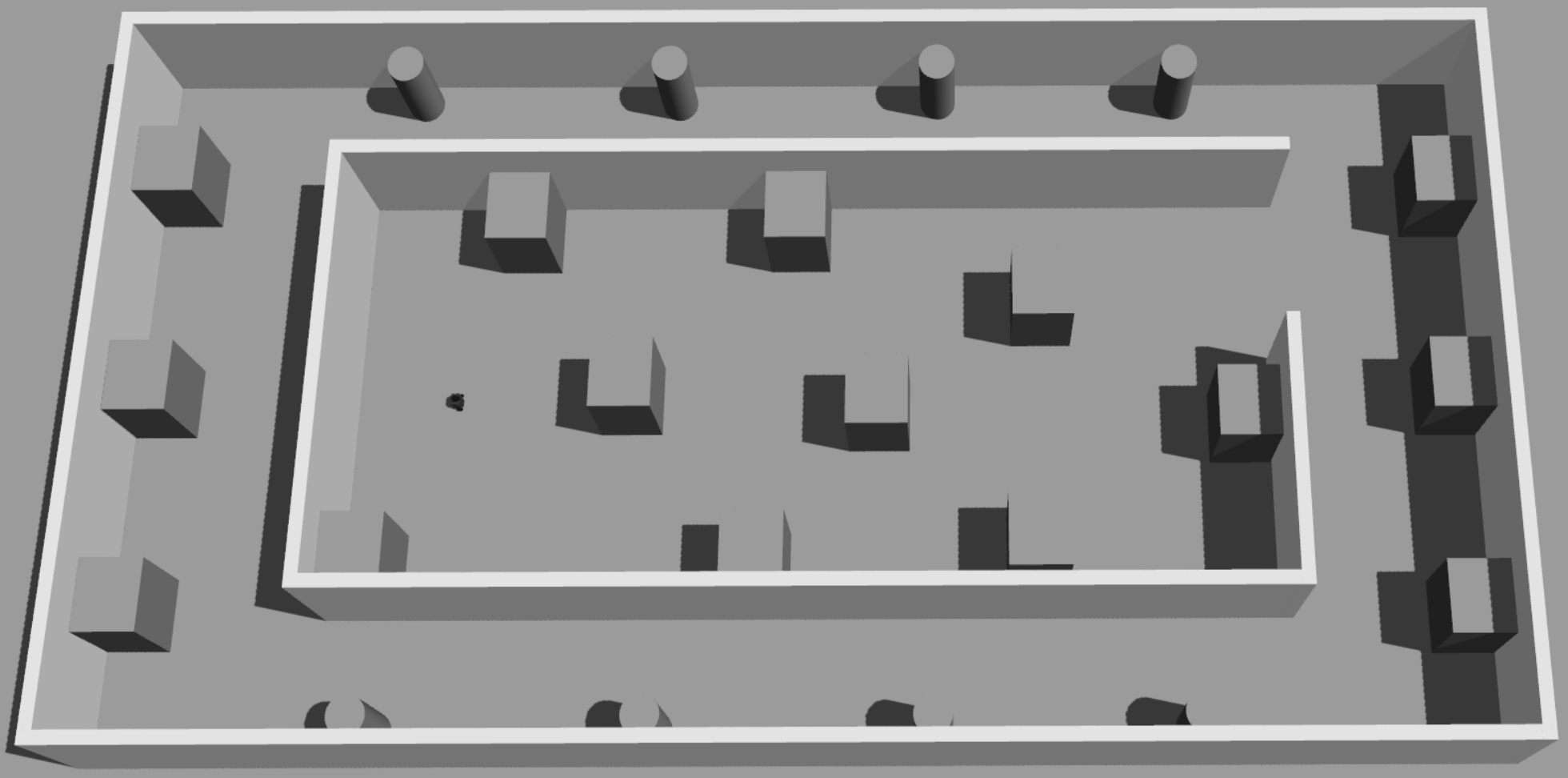}} & Frontier & 778.67 & 60.32 & 99.55\\
		& & RRT & 698.30 & 90.15 & 99.85\\
		& & DA-SLAM & 383.13 & 69.32 & 98.46\\
		& & Ours & \textbf{272.06} & \textbf{52.26} & \textbf{99.93}\\
		\bottomrule
	\end{tabularx}
	\label{table:result}
\end{table}

\begin{figure}
	\centering
	\setlength{\abovecaptionskip}{-0.2cm}
	
	\begin{subfigure}[b]{0.49\textwidth}
		\centering
		\includegraphics[width=\textwidth]{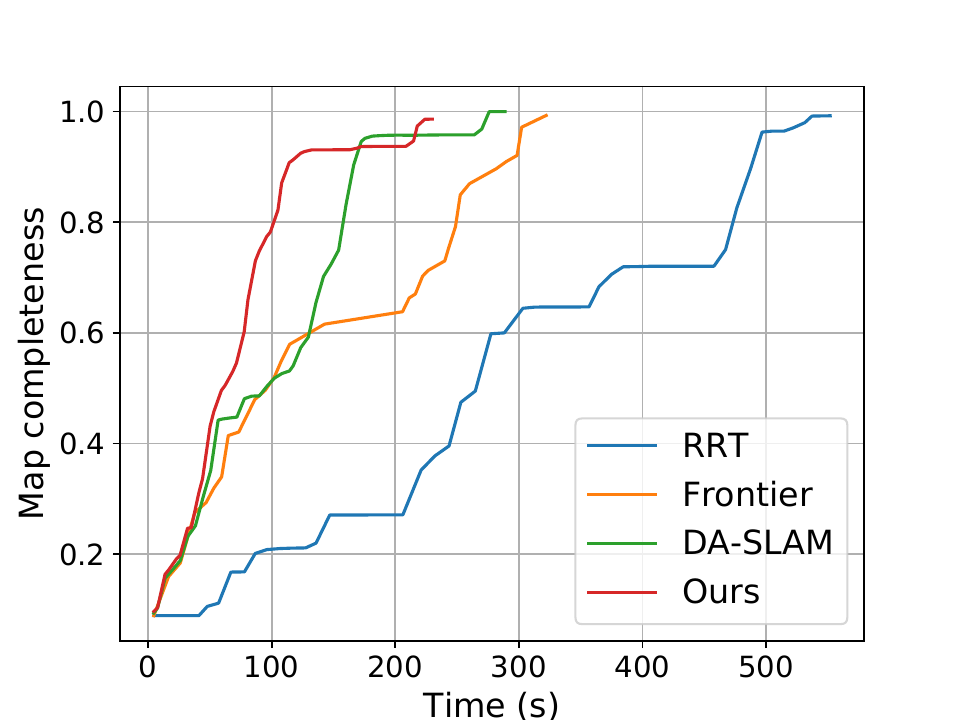}
		\label{fig:5}
	\end{subfigure}
	\hfill
	\begin{subfigure}[b]{0.49\textwidth}
		\centering
		\includegraphics[width=\textwidth]{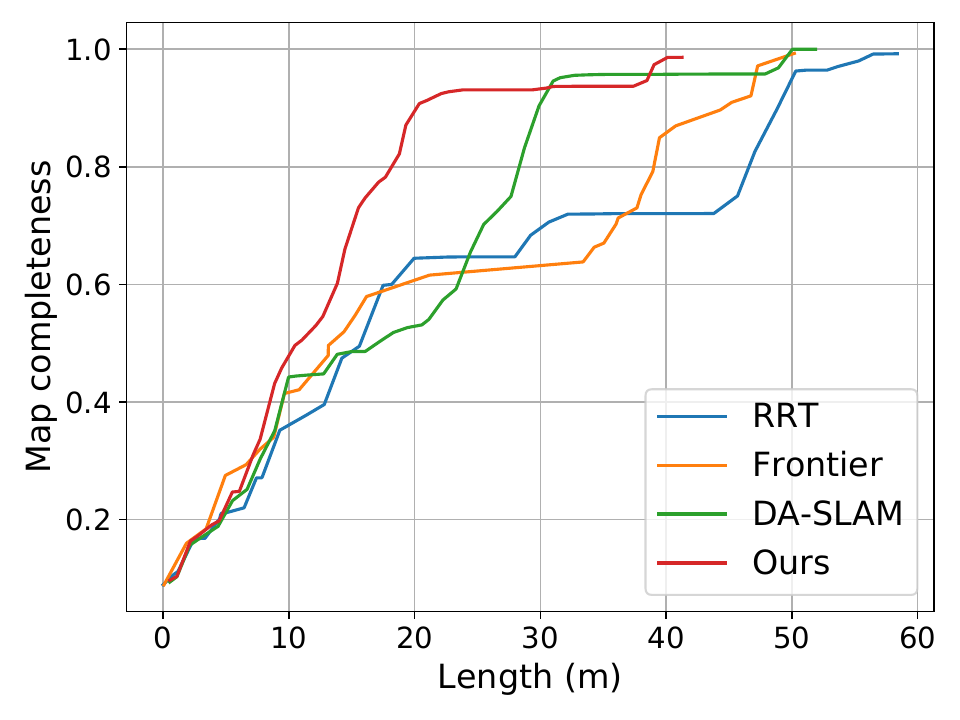}
		\label{fig:6}
	\end{subfigure}
	\vspace{-0.2cm}
	\caption*{(a) test Env-2}
	\vspace{0.8cm}
	\begin{subfigure}[b]{0.49\textwidth}
		\centering
		\includegraphics[width=\textwidth]{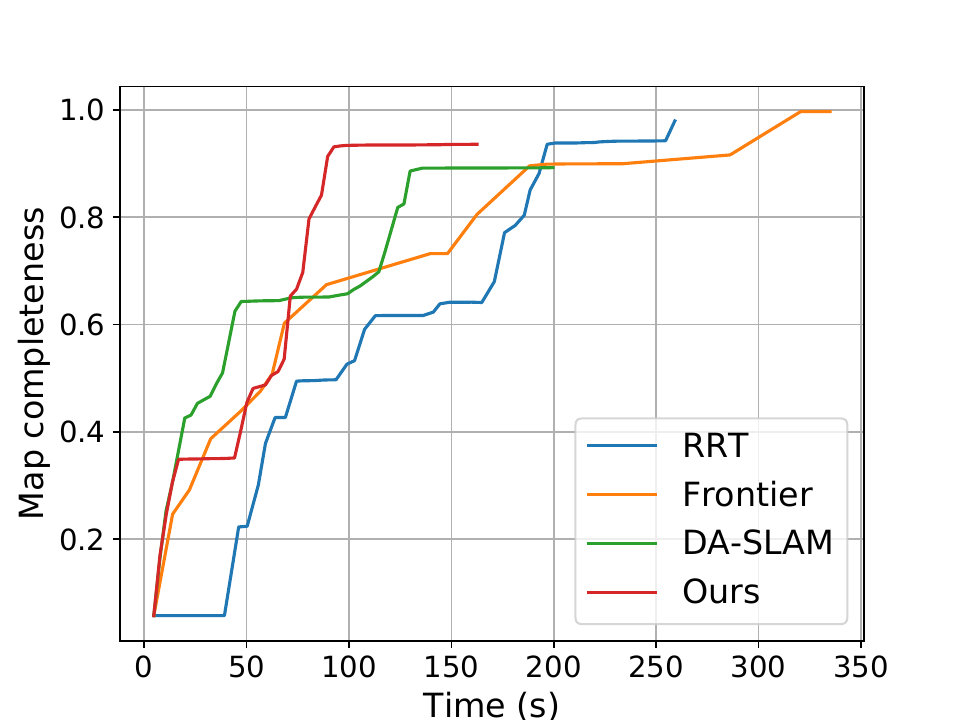}
		\label{fig:7}
	\end{subfigure}
	\hfill
	\begin{subfigure}[b]{0.49\textwidth}
		\centering
		\includegraphics[width=\textwidth]{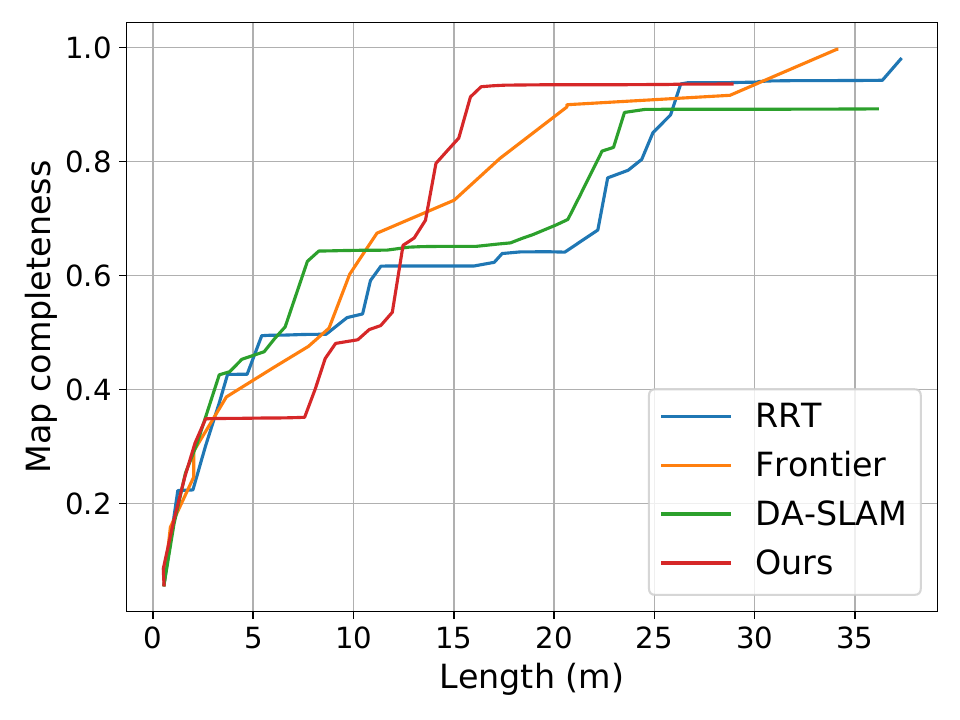}
		\label{fig:8}
	\end{subfigure}
	\vspace{-0.2cm}
	\caption*{(b) test Env-3}
	\vspace{0.8cm}
	\begin{subfigure}[b]{0.49\textwidth}
		\centering
		\includegraphics[width=\textwidth]{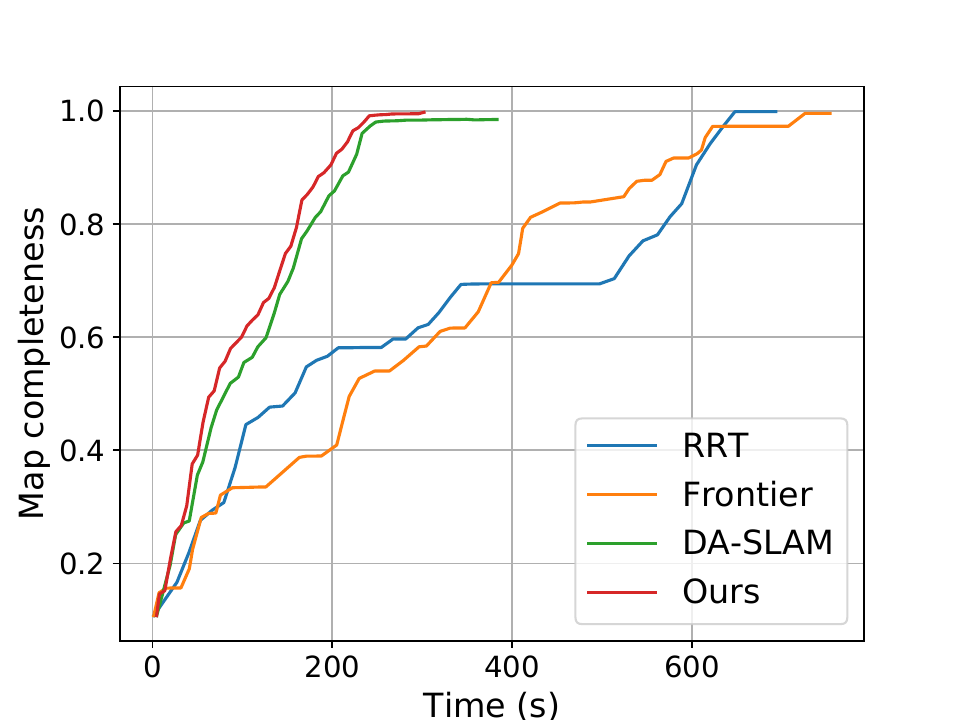}
		\label{fig:3}
	\end{subfigure}
	\hfill
	\begin{subfigure}[b]{0.49\textwidth}
		\centering
		\includegraphics[width=\textwidth]{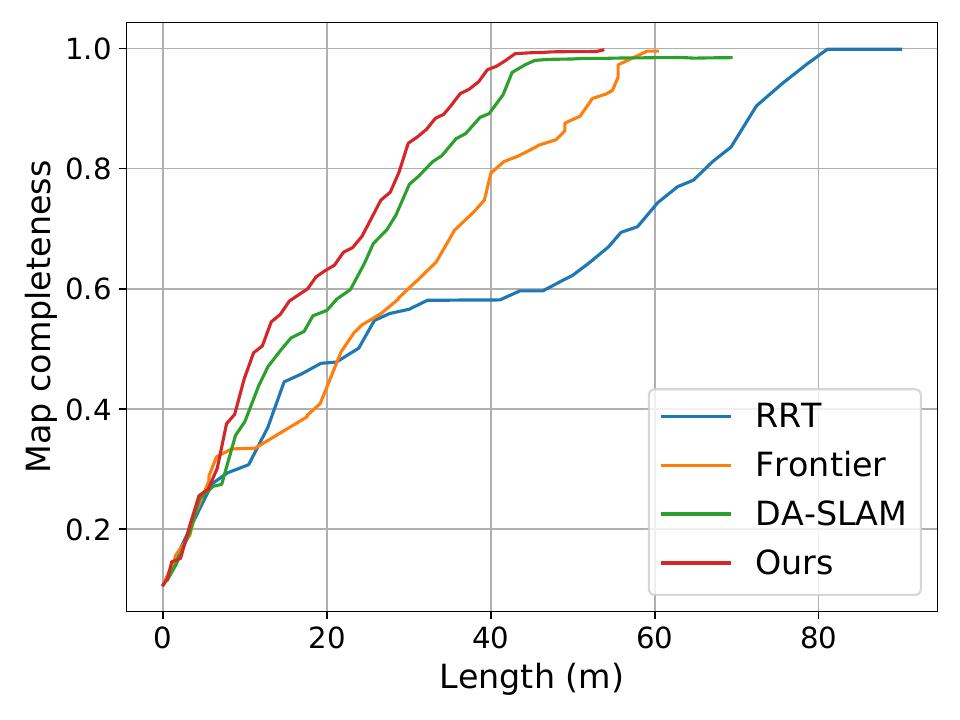}
		\label{fig:4}
	\end{subfigure}
	\vspace{-0.2cm}
	\caption*{(c) test Env-4}
	
	\vspace{0.8cm}
	\caption{Exploration Progress: Coverage vs. Time (Left) and Coverage vs. Path Length (Right).}
	\label{fig:images}
\end{figure}

\subsection{Testing Evaluation}
To comprehensively evaluate the performance of the proposed algorithm, this study designed a systematic comparative experimental framework. In three simulation environments (Env-2, Env-3, and Env-4), the proposed algorithm was compared against Frontier-based\cite{yamauchi1997frontier}, RRT-based\cite{umari2017autonomous} and DA-SLAM (DRL-based)\cite{alcalde2022slam}. Three core metrics were selected: exploration time (from algorithm initiation to automatic termination), robot traversal distance, and map coverage ratio at algorithm termination. Ten independent trials were conducted per scenario to mitigate stochastic effects, with arithmetic means adopted as baseline performance measures. Experimental results are aggregated in Table \ref{table:result}, while \cref{fig:env2,fig:env3,fig:env4} illustrates the exploration paths and mapping outcomes from representative trials (closest to mean performance), and Fig. \ref{fig:images} presents the exploration progress dynamics of each algorithm across different environments.

The results in Table \ref{table:result} show that the proposed algorithm achieves substantial improvements in both exploration time and path length across all test environments. In the most challenging Env-4, the proposed method completes exploration 65$\%$, 60$\%$, and 29$\%$ faster than Frontier, RRT, and DA-SLAM, respectively, while achieving a 42$\%$ shorter path than RRT and a 25$\%$ reduction compared to DA-SLAM. Consistent gains are observed in Env-2 and Env-3, with 25–30$\%$ time savings and approximately 20$\%$ shorter paths. While map completeness is slightly lower than baselines in Env-2 and Env-3, this reflects the algorithm’s deliberate trade-off: by suppressing redundant revisits and prioritizing unexplored regions, it optimizes the balance between efficiency and coverage.
\subsection{Ablation Experiment}
We conducted systematic ablation studies to quantitatively evaluate the contribution of each component in our proposed framework. The experiment was designed with three distinct configurations to isolate the effects of our key innovations:
\begin{enumerate}
	\item Baseline Method: Uncertainty-based DRL approach \cite{alcalde2022slam} (blue curves)
	\item LSD-Enhanced Method: Baseline + Lightweight Stagnation Detection (green curve)
	\item Full Method: Path-Uncertainty Co-Optimization Reward (PUR) + Lightweight Stagnation Detection (red curve)
\end{enumerate}
\begin{figure}
	\centering         
	\includegraphics[width=0.85\textwidth]{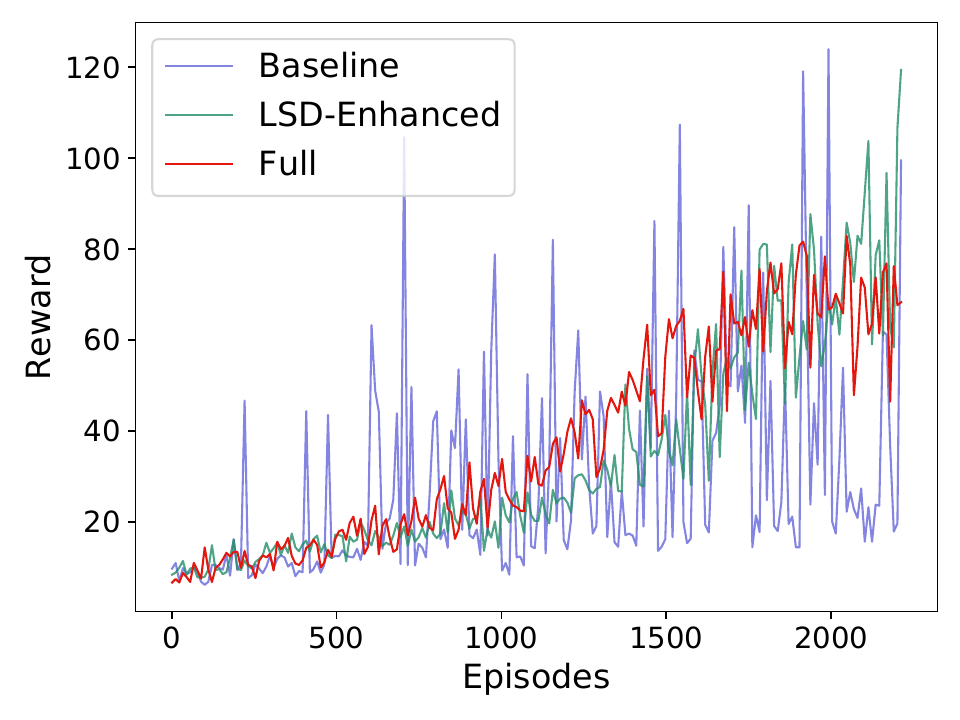}     
	\caption{The average episode rewards during training.}
	\label{fig:reward}
\end{figure}

Figure \ref{fig:reward} presents the average episode rewards over 2000 training episodes. The baseline method demonstrates extremely poor convergence characteristics throughout the entire training process. Despite 2000 episodes of training, its reward curve continues to exhibit severe oscillations, failing to achieve a stable policy. This instability stems from the algorithm's inability to recognize and correct inefficient exploration patterns, allowing the agent to repeatedly learn destructive policies that result in environmental collisions or exploration stagnation. In contrast, both LSD-enhanced variants exhibit significantly stabilized training curves. This stability is attributed to the dual detection mechanism of the LSD module: LiDAR Static Anomaly Detection promptly identifies physical motion failures (such as wheel slippage or collisions), while Map Update Stagnation Detection terminates episodes where exploration progress stagnates despite active movement commands. By terminating these episodes characterized by motion failures or map update stagnation, the mechanism prevents the reinforcement learning agent from incorporating these detrimental experiences into its policy updates. The Full Method not only demonstrates superior stability but also achieves accelerated convergence. The path penalty term (Eq. \ref{eq:2}) actively discourages inefficient exploration behaviors, while the D-optimality criterion ensures sufficient uncertainty reduction for reliable mapping.

\begin{figure}[htbp]
	\centering
	\begin{subfigure}[b]{0.66\textwidth}
		\centering
		\includegraphics[width=\textwidth]{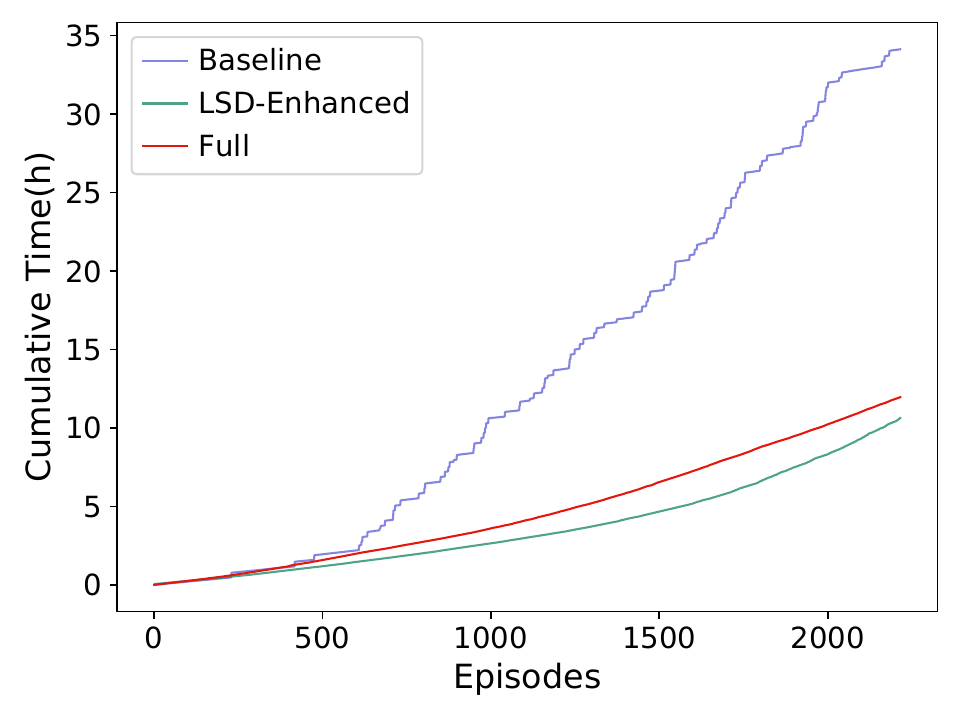}
		\caption{Time vs. Episodes}
		\label{fig:ablation1_time}
	\end{subfigure}
	\begin{subfigure}[b]{0.66\textwidth}
		\centering
		\includegraphics[width=\textwidth]{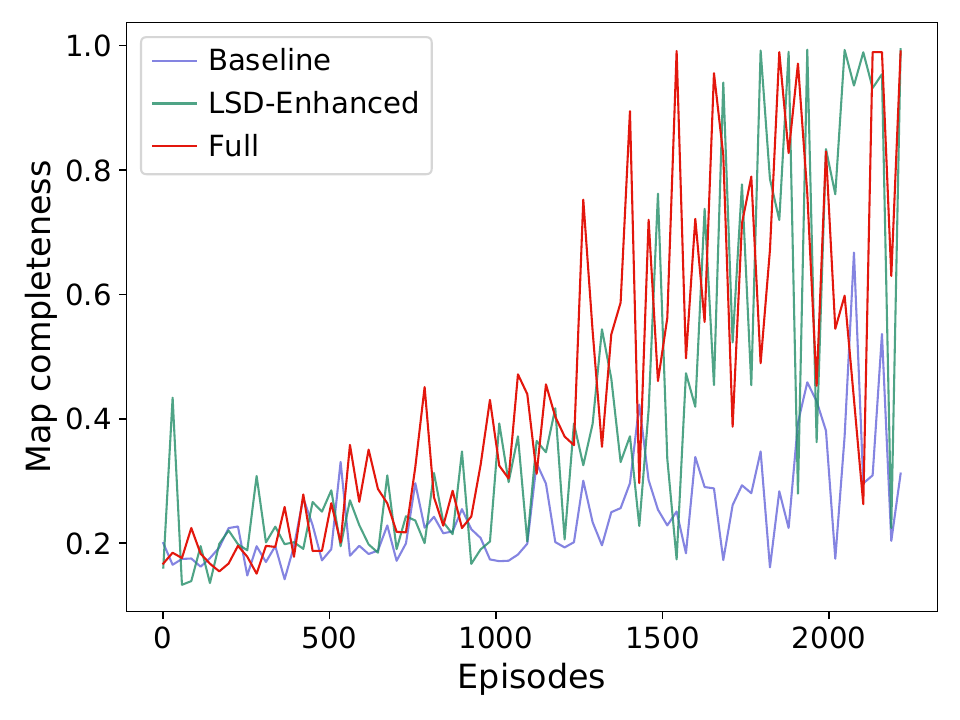}
		\caption{Coverage vs. Episodes}
		\label{fig:ablation1_ratio}
	\end{subfigure}
	\begin{subfigure}[b]{0.66\textwidth}
		\centering
		\includegraphics[width=\textwidth]{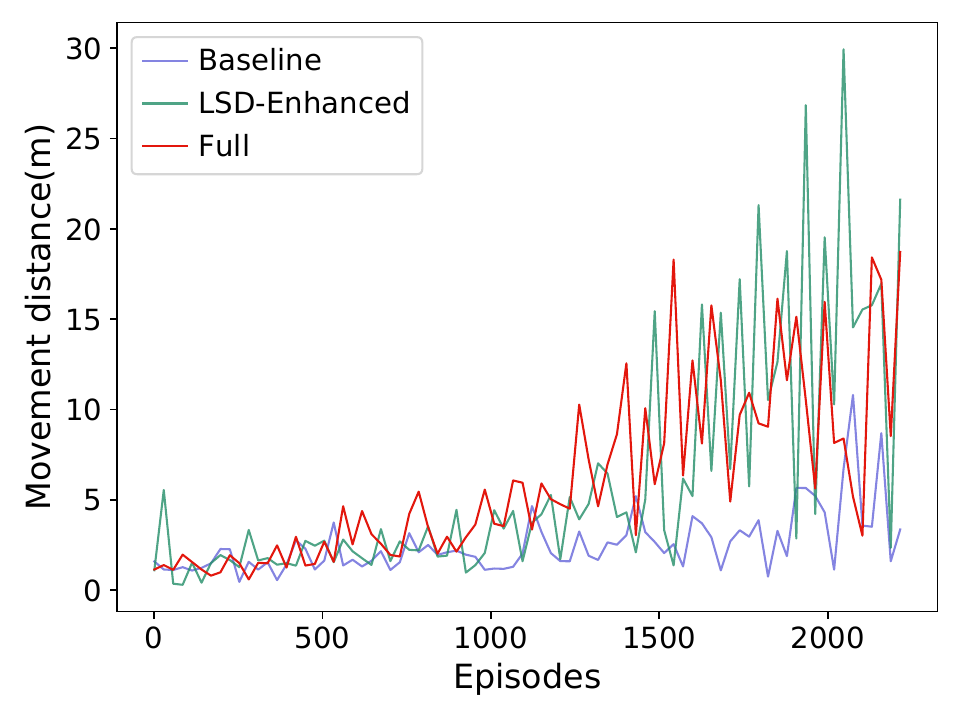}
		\caption{Distance vs. Episodes}
		\label{fig:ablation1_length}
	\end{subfigure}
	\caption{Effectiveness of LSD and PUR during training.}
	\label{fig:ablation1}
\end{figure}

Fig. \ref{fig:ablation1} presents a comprehensive analysis of our ablation study across three critical performance metrics. Regarding exploration time, both the LSD-enhanced and the Full methods significantly outperform the baseline. Notably, the LSD-enhanced method exhibits shorter exploration time than the Full method. This apparent discrepancy is explained by analyzing map completeness (Fig. \ref{fig:ablation1_ratio}) and movement distance (Fig. \ref{fig:ablation1_length}): The Full method achieves faster convergence, attaining higher map completeness within the same training episodes (the red curve in Fig. \ref{fig:ablation1_ratio} reaches high coverage earlier). This thorough exploration behavior increases exploration time and movement distance (higher distance values for the red curve in Fig. \ref{fig:ablation1_length}). In contrast, while the LSD-enhanced method optimizes stagnation issues, its lack of dynamic reward guidance limits exploration scope within the same episodes, resulting in shorter time but incomplete map coverage. The baseline method performs worst across all metrics, exhibiting substantial time waste and redundant movement, underscoring the necessity of action correction and reward optimization.

This analysis validates the synergistic relationship between our two proposed mechanisms: the LSD module provides essential execution-level stability by filtering out catastrophic failures, while the DRL framework optimizes decision-making efficiency through a balanced reward structure. Their integration creates a learning environment where policy updates are derived from consistently productive exploration experiences, thereby ensuring training stability and accelerating convergence speed.
\subsection{Real-world Experiment}
To validate the effectiveness and adaptability of the proposed method in real-world scenarios, this study further conducted physical platform experiments. The experimental environment was set up within our laboratory, covering an area of approximately 60 square meters, with numerous randomly placed obstacles including chairs, tables, storage cabinets, and cardboard boxes. This environmental layout simulates the common cluttered obstacle distribution found in practical applications, presenting significant challenges to the robustness and adaptability of exploration algorithms. The experimental platform was equipped with an Intel N100 processor (1.8 GHz clock speed, 4 cores and 4 threads) running Ubuntu 20.04 operating system, and integrated with a Hokuyo UST-10LX LiDAR (ranging distance of 0.1-3.5 meters). To ensure experimental safety, the robot's maximum linear velocity was set to 0.15 m/s, slightly lower than that used in the simulation environment.

\begin{figure}
	\centering         
	\includegraphics[width=0.85\textwidth]{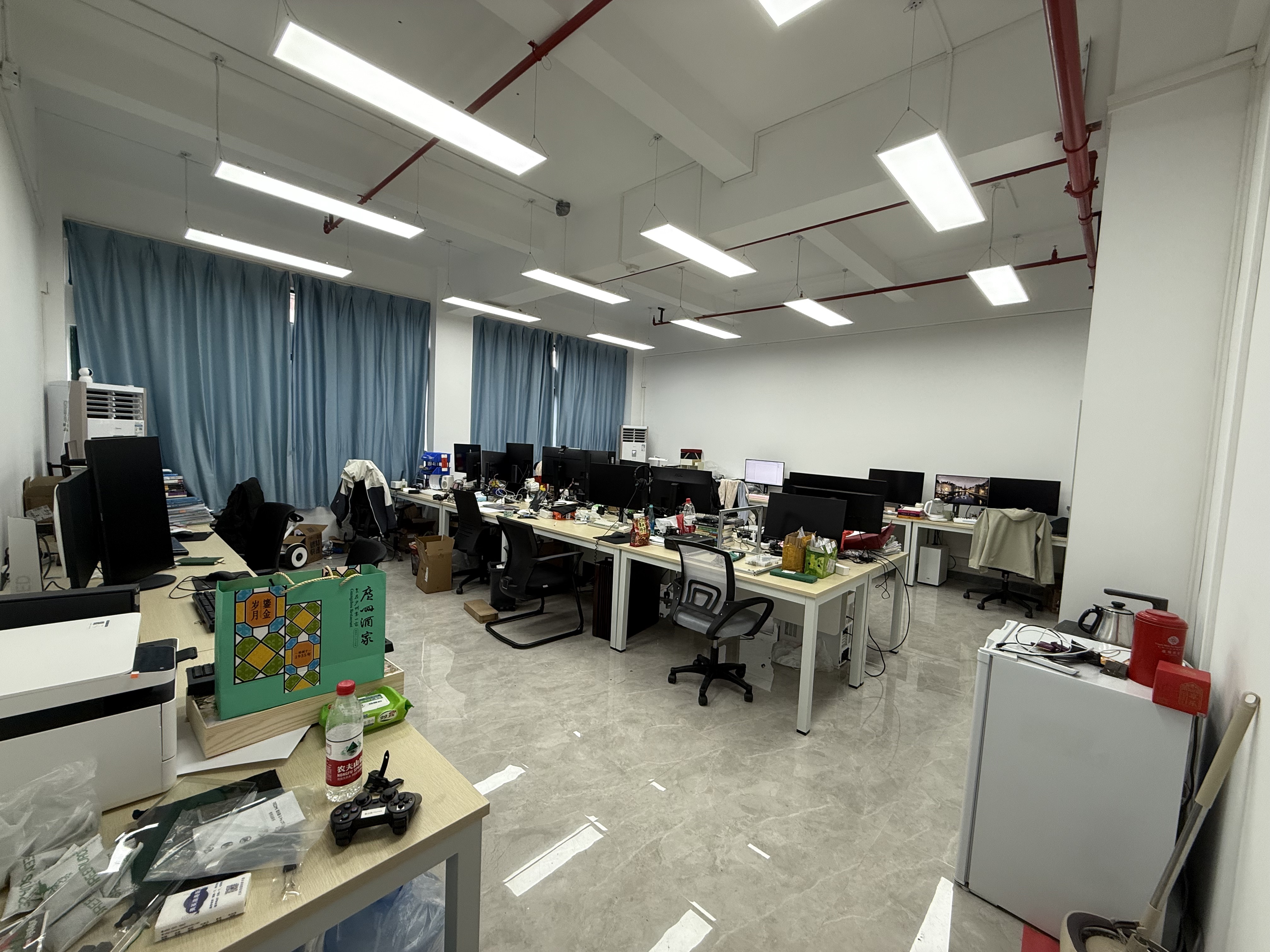}     
	\caption{Laboratory environment.}
	\label{fig:realworld}
\end{figure}

\begin{figure}
	\centering         
	\includegraphics[width=0.4\textwidth]{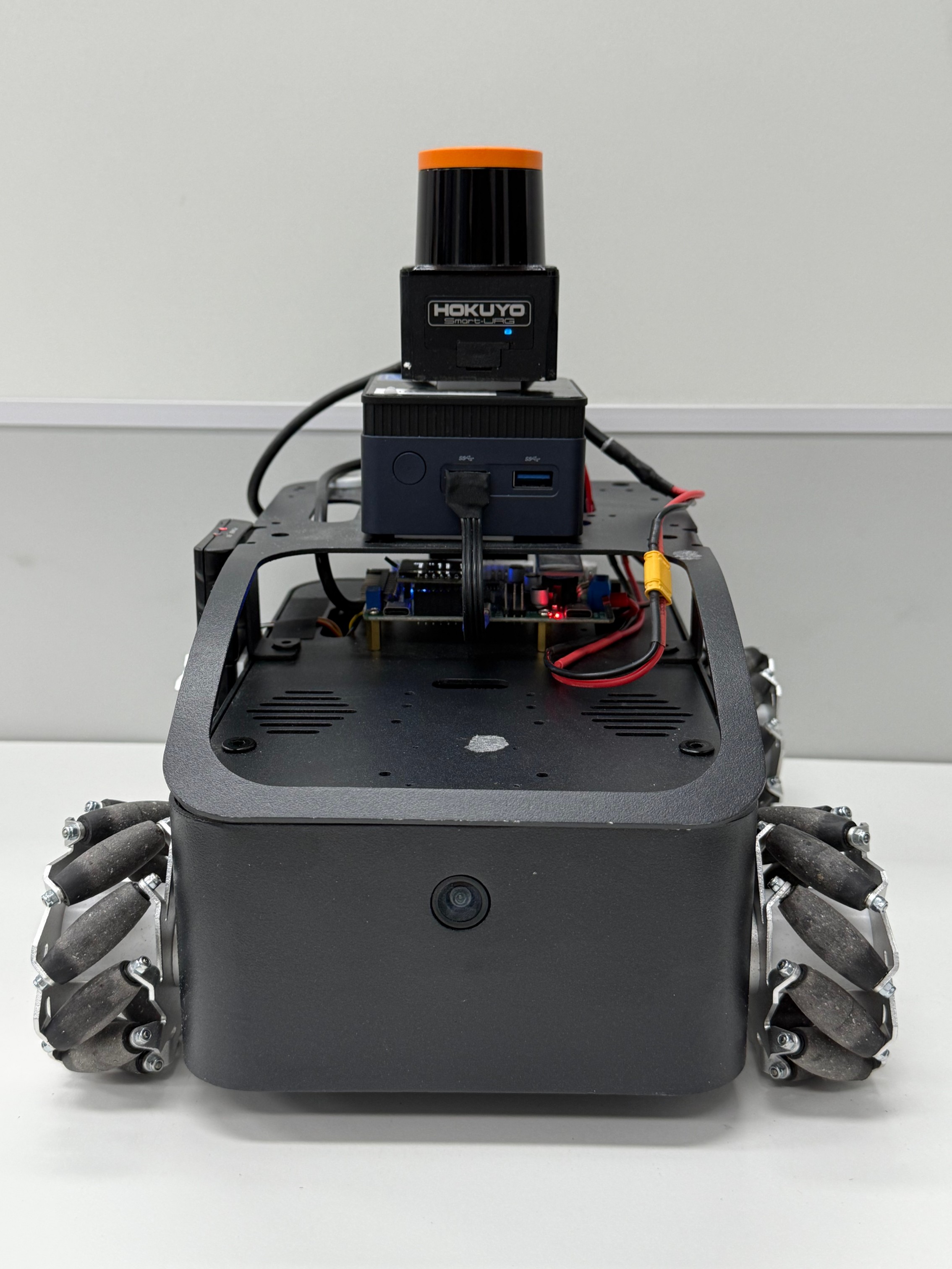}     
	\caption{Experimental platform}
	\label{fig:realrobot}
\end{figure}

\begin{figure}
	\centering         
	\includegraphics[width=0.75\textwidth]{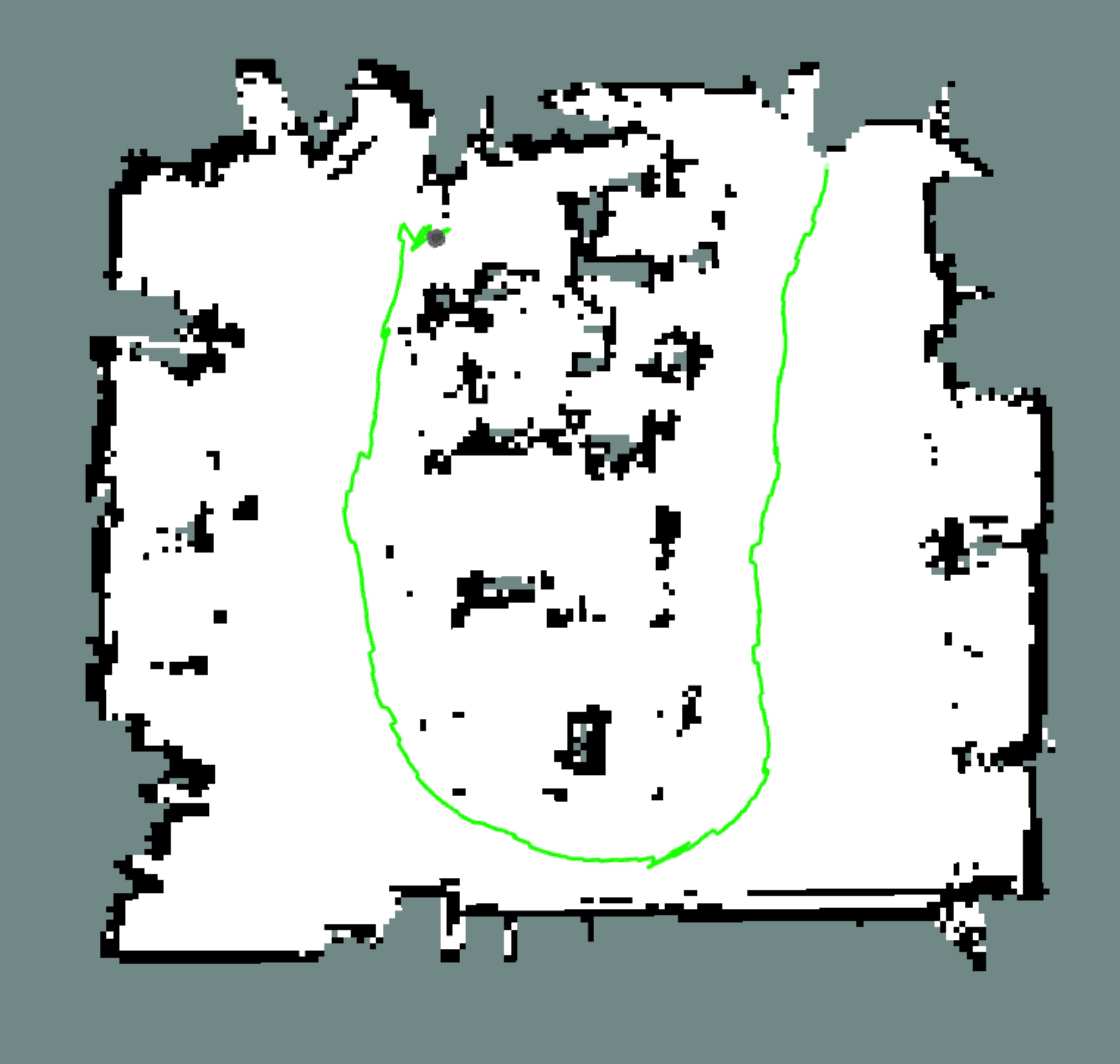}     
	\caption{The constructed map}
	\label{fig:realmap}
\end{figure}

Fig. \ref{fig:realworld} shows the laboratory environment, clearly displaying the densely distributed obstacles and complex spatial structures. Fig. \ref{fig:realrobot} illustrates the experimental platform. Fig. \ref{fig:realmap} presents the robot's final exploration trajectory (green path) along with the constructed complete environmental map. The experimental results indicate that the robot completed the entire environment exploration within 122.32 seconds, with a final path length of 15.96 meters, while successfully avoiding collisions with various obstacles in the environment. This result shows high consistency with the performance observed in the simulation environment, validating the transferability of the proposed algorithm from simulation to real-world settings. Notably, the robot's exploration path during the process exhibited a highly concentrated characteristic, significantly reducing unnecessary backtracking and detouring behaviors.

The real-world experimental results verify the reliability of the simulation experiments and demonstrate the practical value of our proposed method, laying the foundation for future deployment in real-world applications such as disaster rescue and underground mine exploration. In our subsequent research, we will focus on validating the algorithm's performance in larger-scale and more dynamic real environments.

\section{CONCLUSIONS}
This paper proposes a Path-Uncertainty Co-Optimization DRL framework with Lightweight Stagnation Detection mechanism, which significantly enhances robotic exploration efficiency. Experimental results show that the proposed method substantially outperforms the frontier-based method, the RRT-based approach and DA-SLAM in terms of exploration efficiency, while maintaining reliable map completeness. Specifically, it reduces exploration time by up to 65$\%$ relative to the frontier-based method and shortens path length by up to 42$\%$ compared to RRT, consistently outperforming all baselines across diverse and complex scenarios. Ablation studies further confirm the complementary roles of the Path-Uncertainty Co-Optimization Reward and the Lightweight Stagnation Detection mechanism: PUR enhances the efficient allocation of exploration resources at the decision-making level, while LSD suppresses ineffective behaviors at the execution level. Their synergistic integration significantly accelerates training convergence. Moreover, real-world experiments on a physical robotic platform validate the successful sim-to-real transferability of the proposed approach, demonstrating its practical applicability and deployment potential. Future work will focus on further optimizing coverage uniformity and enhancing system robustness in large-scale environments.

%
% Each of the commands below will create an unnumbered section with the appropriate heading.
% Remove any sections that are not relevant for your article.
% All sections except suppdata will be removed if the [anonymous] option is used.
% See iopjournal-guidelines.pdf for more information.
%

%\section*{References}
\bibliographystyle{IEEEtran}
\bibliography{ref}

@article{placed2023survey,
	title={A survey on active simultaneous localization and mapping: State of the art and new frontiers},
	author={Placed, Julio A and Strader, Jared and Carrillo, Henry and Atanasov, Nikolay and Indelman, Vadim and Carlone, Luca and Castellanos, Jos{\'e} A},
	journal={IEEE Transactions on Robotics},
	volume={39},
	number={3},
	pages={1686--1705},
	year={2023},
	publisher={IEEE}
}

@article{niroui2019deep,
	title={Deep reinforcement learning robot for search and rescue applications: Exploration in unknown cluttered environments},
	author={Niroui, Farzad and Zhang, Kaicheng and Kashino, Zendai and Nejat, Goldie},
	journal={IEEE Robotics and Automation Letters},
	volume={4},
	number={2},
	pages={610--617},
	year={2019},
	publisher={IEEE}
}

@article{nahavandi2025comprehensive,
	title={A comprehensive review on autonomous navigation},
	author={Nahavandi, Saeid and Alizadehsani, Roohallah and Nahavandi, Darius and Mohamed, Shady and Mohajer, Navid and Rokonuzzaman, Mohammad and Hossain, Ibrahim},
	journal={ACM Computing Surveys},
	volume={57},
	number={9},
	pages={1--67},
	year={2025},
	publisher={ACM New York, NY}
}

@article{wang2025multi,
	title={Multi-Robot System for Cooperative Exploration in Unknown Environments: A Survey},
	author={Wang, Chuqi and Yu, Chao and Xu, Xin and Gao, Yuman and Yang, Xinyi and Tang, Wenhao and Yu, Shu'ang and Chen, Yinuo and Gao, Feng and Jian, ZhuoZhu and others},
	journal={arXiv preprint arXiv:2503.07278},
	year={2025}
}

@article{ackerman2022robots,
	title={Robots conquer the underground: What DARPA's subterranean challenge means for the future of autonomous robots},
	author={Ackerman, Evan},
	journal={IEEE Spectrum},
	volume={59},
	number={5},
	pages={30--37},
	year={2022},
	publisher={IEEE}
}

@article{zhou2023racer,
	title={Racer: Rapid collaborative exploration with a decentralized multi-uav system},
	author={Zhou, Boyu and Xu, Hao and Shen, Shaojie},
	journal={IEEE Transactions on Robotics},
	volume={39},
	number={3},
	pages={1816--1835},
	year={2023},
	publisher={IEEE}
}

@article{azpurua2023survey,
	title={A Survey on the autonomous exploration of confined subterranean spaces: Perspectives from real-word and industrial robotic deployments},
	author={Azp{\'u}rua, H{\'e}ctor and Saboia, Ma{\'\i}ra and Freitas, Gustavo M and Clark, Lillian and Agha-mohammadi, Ali-akbar and Pessin, Gustavo and Campos, Mario FM and Macharet, Douglas G},
	journal={Robotics and Autonomous Systems},
	volume={160},
	pages={104304},
	year={2023},
	publisher={Elsevier}
}

@inproceedings{yamauchi1997frontier,
	title={A frontier-based approach for autonomous exploration},
	author={Yamauchi, Brian},
	booktitle={Proceedings 1997 IEEE International Symposium on Computational Intelligence in Robotics and Automation CIRA'97.'Towards New Computational Principles for Robotics and Automation'},
	pages={146--151},
	year={1997},
	organization={IEEE}
}

@article{keidar2014efficient,
	title={Efficient frontier detection for robot exploration},
	author={Keidar, Matan and Kaminka, Gal A},
	journal={The International Journal of Robotics Research},
	volume={33},
	number={2},
	pages={215--236},
	year={2014},
	publisher={SAGE Publications Sage UK: London, England}
}

@article{wang2024exploration,
	title={An exploration-enhanced search algorithm for robot indoor source searching},
	author={Wang, Miao and Xin, Bin and Jing, Mengjie and Qu, Yun},
	journal={IEEE Transactions on Robotics},
	year={2024},
	publisher={IEEE}
}

@article{saravanan2024fit,
	title={FIT-SLAM--Fisher Information and Traversability estimation-based Active SLAM for exploration in 3D environments},
	author={Saravanan, Suchetan and Chauffaut, Corentin and Chanel, Caroline and Vivet, Damien},
	journal={arXiv preprint arXiv:2401.09322},
	year={2024}
}

@inproceedings{umari2017autonomous,
	title={Autonomous robotic exploration based on multiple rapidly-exploring randomized trees},
	author={Umari, Hassan and Mukhopadhyay, Shayok},
	booktitle={2017 IEEE/RSJ International Conference on Intelligent Robots and Systems (IROS)},
	pages={1396--1402},
	year={2017},
	organization={IEEE}
}

@inproceedings{faust2018prm,
	title={Prm-rl: Long-range robotic navigation tasks by combining reinforcement learning and sampling-based planning},
	author={Faust, Aleksandra and Oslund, Kenneth and Ramirez, Oscar and Francis, Anthony and Tapia, Lydia and Fiser, Marek and Davidson, James},
	booktitle={2018 IEEE international conference on robotics and automation (ICRA)},
	pages={5113--5120},
	year={2018},
	organization={IEEE}
}

@inproceedings{wu2019autonomous,
	title={Autonomous mobile robot exploration in unknown indoor environments based on rapidly-exploring random tree},
	author={Wu, Cheng-Yan and Lin, Huei-Yung},
	booktitle={2019 IEEE International Conference on Industrial Technology (ICIT)},
	pages={1345--1350},
	year={2019},
	organization={IEEE}
}

@article{whaite1997autonomous,
	title={Autonomous exploration: Driven by uncertainty},
	author={Whaite, Peter and Ferrie, Frank P},
	journal={IEEE Transactions on Pattern Analysis and Machine Intelligence},
	volume={19},
	number={3},
	pages={193--205},
	year={1997},
	publisher={IEEE}
}

@inproceedings{stachniss2005information,
	title={Information gain-based exploration using rao-blackwellized particle filters.},
	author={Stachniss, Cyrill and Grisetti, Giorgio and Burgard, Wolfram},
	booktitle={Robotics: Science and systems},
	volume={2},
	number={1},
	pages={65--72},
	year={2005}
}

@inproceedings{carrillo2012comparison,
	title={On the comparison of uncertainty criteria for active SLAM},
	author={Carrillo, Henry and Reid, Ian and Castellanos, Jos{\'e} A},
	booktitle={2012 IEEE International Conference on Robotics and Automation},
	pages={2080--2087},
	year={2012},
	organization={IEEE}
}

@article{zhao2024autonomous,
	title={Autonomous driving system: A comprehensive survey},
	author={Zhao, Jingyuan and Zhao, Wenyi and Deng, Bo and Wang, Zhenghong and Zhang, Feng and Zheng, Wenxiang and Cao, Wanke and Nan, Jinrui and Lian, Yubo and Burke, Andrew F},
	journal={Expert Systems with Applications},
	volume={242},
	pages={122836},
	year={2024},
	publisher={Elsevier}
}

@article{asgharivaskasi2023semantic,
	title={Semantic octree mapping and shannon mutual information computation for robot exploration},
	author={Asgharivaskasi, Arash and Atanasov, Nikolay},
	journal={IEEE Transactions on Robotics},
	volume={39},
	number={3},
	pages={1910--1928},
	year={2023},
	publisher={IEEE}
}

@article{zhou2024indoor,
	title={An indoor blind area-oriented autonomous robotic path planning approach using deep reinforcement learning},
	author={Zhou, Yuting and Yang, Junchao and Guo, Zhiwei and Shen, Yu and Yu, Keping and Lin, Jerry Chun-Wei},
	journal={Expert Systems with Applications},
	volume={254},
	pages={124277},
	year={2024},
	publisher={Elsevier}
}

@article{zhao2024exploration,
	title={Exploration-and Exploitation-Driven Deep Deterministic Policy Gradient for Active SLAM in Unknown Indoor Environments},
	author={Zhao, Shengmin and Hwang, Seung-Hoon},
	journal={Electronics},
	volume={13},
	number={5},
	pages={999},
	year={2024},
	publisher={MDPI}
}

@article{cao2023ariadne,
	title={Ariadne: A reinforcement learning approach using attention-based deep networks for exploration},
	author={Cao, Yuhong and Hou, Tianxiang and Wang, Yizhuo and Yi, Xian and Sartoretti, Guillaume},
	journal={arXiv preprint arXiv:2301.11575},
	year={2023}
}

@article{chen2024lidar,
	title={LiDAR-based end-to-end active SLAM using deep reinforcement learning in large-scale environments},
	author={Chen, Jiaying and Wu, Keyu and Hu, Minghui and Suganthan, Ponnuthurai Nagaratnam and Makur, Anamitra},
	journal={IEEE Transactions on Vehicular Technology},
	year={2024},
	publisher={IEEE}
}

@article{botteghi2021curiosity,
	title={Curiosity-driven reinforcement learning agent for mapping unknown indoor environments},
	author={Botteghi, Nicol{\`o} and Sirmacek, B and Poel, M and Brune, C and Schulte, R},
	journal={ISPRS Annals of the Photogrammetry, Remote Sensing and Spatial Information Sciences},
	volume={5},
	number={1},
	pages={129--136},
	year={2021},
	publisher={Copernicus}
}

@inproceedings{alcalde2022slam,
	title={DA-SLAM: Deep active SLAM based on deep reinforcement learning},
	author={Alcalde, Martin and Ferreira, Matias and Gonz{\'a}lez, Pablo and Andrade, Federico and Tejera, Gonzalo},
	booktitle={2022 Latin American Robotics Symposium (LARS), 2022 Brazilian Symposium on Robotics (SBR), and 2022 Workshop on Robotics in Education (WRE)},
	pages={282--287},
	year={2022},
	organization={IEEE}
}

@article{placed2020deep,
	title={A deep reinforcement learning approach for active SLAM},
	author={Placed, Julio A and Castellanos, Jos{\'e} A},
	journal={Applied Sciences},
	volume={10},
	number={23},
	pages={8386},
	year={2020},
	publisher={MDPI}
}

@inproceedings{zhu2018deep,
	title={Deep reinforcement learning supervised autonomous exploration in office environments},
	author={Zhu, Delong and Li, Tingguang and Ho, Danny and Wang, Chaoqun and Meng, Max Q-H},
	booktitle={2018 IEEE international conference on robotics and automation (ICRA)},
	pages={7548--7555},
	year={2018},
	organization={IEEE}
}

@article{chaplot2020learning,
	title={Learning to explore using active neural slam},
	author={Chaplot, Devendra Singh and Gandhi, Dhiraj and Gupta, Saurabh and Gupta, Abhinav and Salakhutdinov, Ruslan},
	journal={arXiv preprint arXiv:2004.05155},
	year={2020}
}

@article{zhao2025multirobot,
	title={Multirobot unknown environment exploration and obstacle avoidance based on a Voronoi diagram and reinforcement learning},
	author={Zhao, Hongyang and Guo, Yanan and Liu, Yi and Jin, Jing},
	journal={Expert Systems with Applications},
	volume={264},
	pages={125900},
	year={2025},
	publisher={Elsevier}
}

@article{schulman2017proximal,
	title={Proximal policy optimization algorithms},
	author={Schulman, John and Wolski, Filip and Dhariwal, Prafulla and Radford, Alec and Klimov, Oleg},
	journal={arXiv preprint arXiv:1707.06347},
	year={2017}
}

@inproceedings{quigley2009ros,
	title={ROS: an open-source Robot Operating System},
	author={Quigley, Morgan and Conley, Ken and Gerkey, Brian and Faust, Josh and Foote, Tully and Leibs, Jeremy and Wheeler, Rob and Ng, Andrew Y and others},
	booktitle={ICRA workshop on open source software},
	volume={3},
	number={3.2},
	pages={5},
	year={2009},
	organization={Kobe}
}

\end{document}